\documentclass[10pt,journal,final]{IEEEtran}
\usepackage[utf8]{inputenc}
\usepackage{cite}
\usepackage{CJK}
\usepackage{tipa}
\usepackage{amsmath}
\usepackage{enumerate}
\usepackage{graphicx}
\graphicspath{{figure/}}
\usepackage{multirow}
\usepackage{array}
\usepackage{stfloats}
\usepackage{multicol}
\usepackage{amsthm,amsmath,amssymb}
\usepackage{mathrsfs}
\usepackage{amssymb}
\usepackage{booktabs}
\usepackage[normalem]{ulem} 
\usepackage{subfigure}

\usepackage[ruled]{algorithm2e}

\usepackage{color}
\definecolor{dgreen}{rgb}{0,.6,0}

\usepackage{soul}

\begin{document}
\title{Detecting Recolored Image by Spatial Correlation}
\author
{\IEEEauthorblockN{Yushu Zhang},
        {Nuo Chen},
        {Shuren Qi}, 
        {Mingfu Xue},
        and
        {Xiaochun Cao}

\thanks{Y. Zhang, N. Chen, S. Qi and M. Xue are with the College of Computer Science and Technology, Nanjing University of Aeronautics and Astronautics, Nanjing 210016, China (e-mails: yushu@nuaa.edu.cn; yournuo@nuaa.edu.cn; shurenqi@nuaa.edu.cn; mingfu.xue@nuaa.edu.cn). }
\thanks{X. Cao is with the School of Cyber Science and Technology, Sun Yat-sen University, Shenzhen, China (e-mail:
caoxch5@sysu.edu.cn) }
}

\maketitle
\begin{abstract}
Image forensics, aiming to ensure the authenticity of the image, has made great progress in dealing with common image manipulation such as copy-move, splicing, and inpainting in the past decades. However, only a few researchers pay attention to an emerging editing technique called image recoloring, which can manipulate the color values of an image to give it a new style. To prevent it from being used maliciously, the previous approaches address the conventional recoloring from the perspective of inter-channel correlation and illumination consistency. In this paper, we try to explore a solution from the perspective of the spatial correlation, which exhibits the generic detection capability for both conventional and deep learning-based recoloring. Through theoretical and numerical analysis, we find that the recoloring operation will inevitably destroy the spatial correlation between pixels, implying a new prior of statistical discriminability. Based on such fact, we generate a set of spatial correlation features and learn the informative representation from the set via a convolutional neural network. To train our network, we use three recoloring methods to generate a large-scale and high-quality data set. Extensive experimental results in two recoloring scenes demonstrate that the spatial correlation features are highly discriminative. Our method achieves the state-of-the-art detection accuracy on multiple benchmark datasets and exhibits well generalization for unknown types of recoloring methods.
\end{abstract}


\IEEEpeerreviewmaketitle

\section{Introduction}
As an important information medium, images are widely used to distribute news and record events. For a long time, people are accustomed to \textit{seeing is believing}. However, with the rapid proliferation of image editing software such as Adobe Photoshop and GIMP, anyone can easily edit digital images at a very low cost, and the generated content is quite difficult to distinguish for humans \cite{nightingale2017can}. In addition to the most common editing operations, such as copy-move, splicing, and inpainting, a new type of editing technique, image recoloring, has emerged.

\begin{figure}[htbp]
\centering
\subfigure[]{
\includegraphics[scale=0.29]{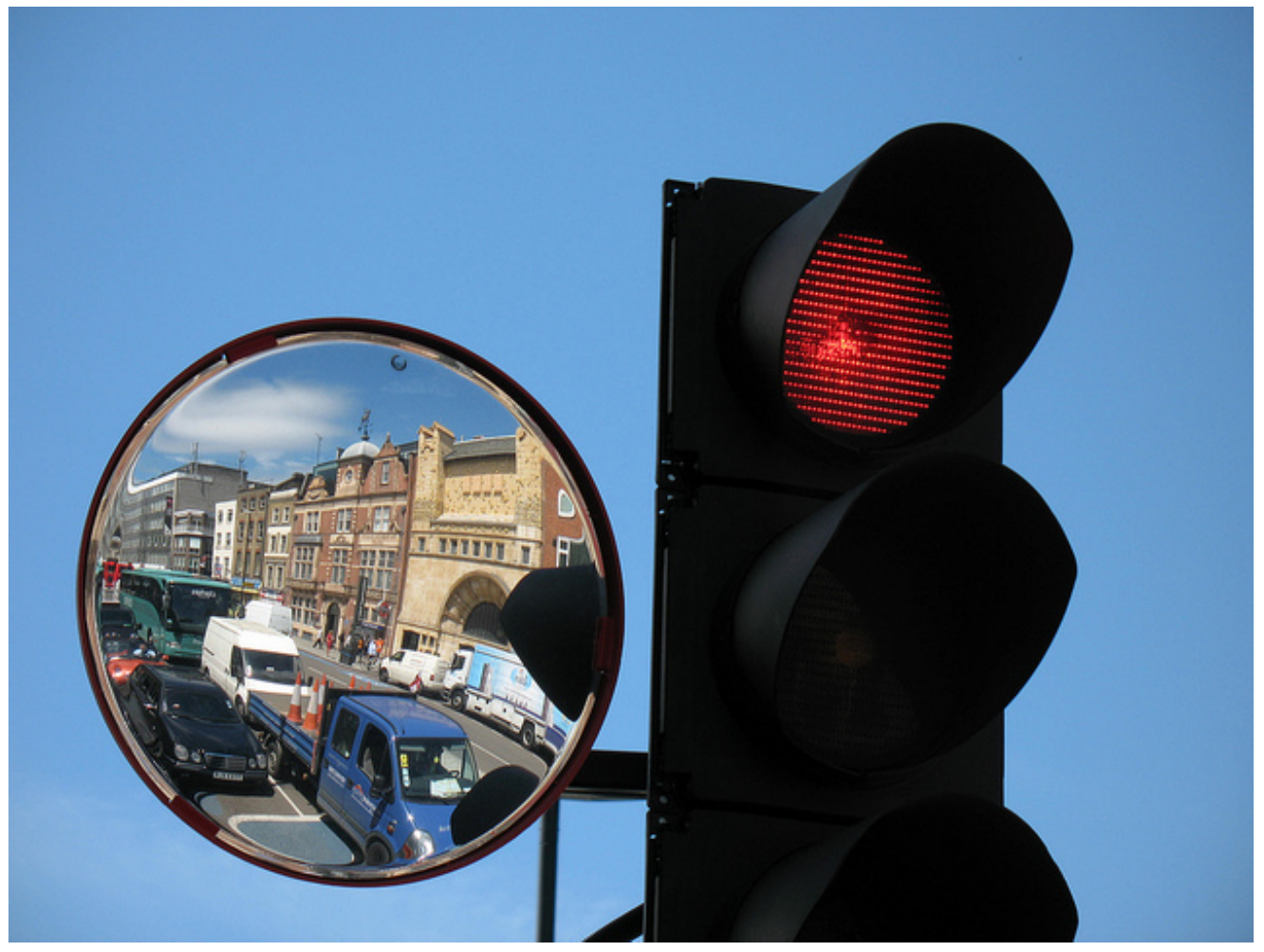}
}
\quad
\subfigure[]{
\includegraphics[scale=0.29]{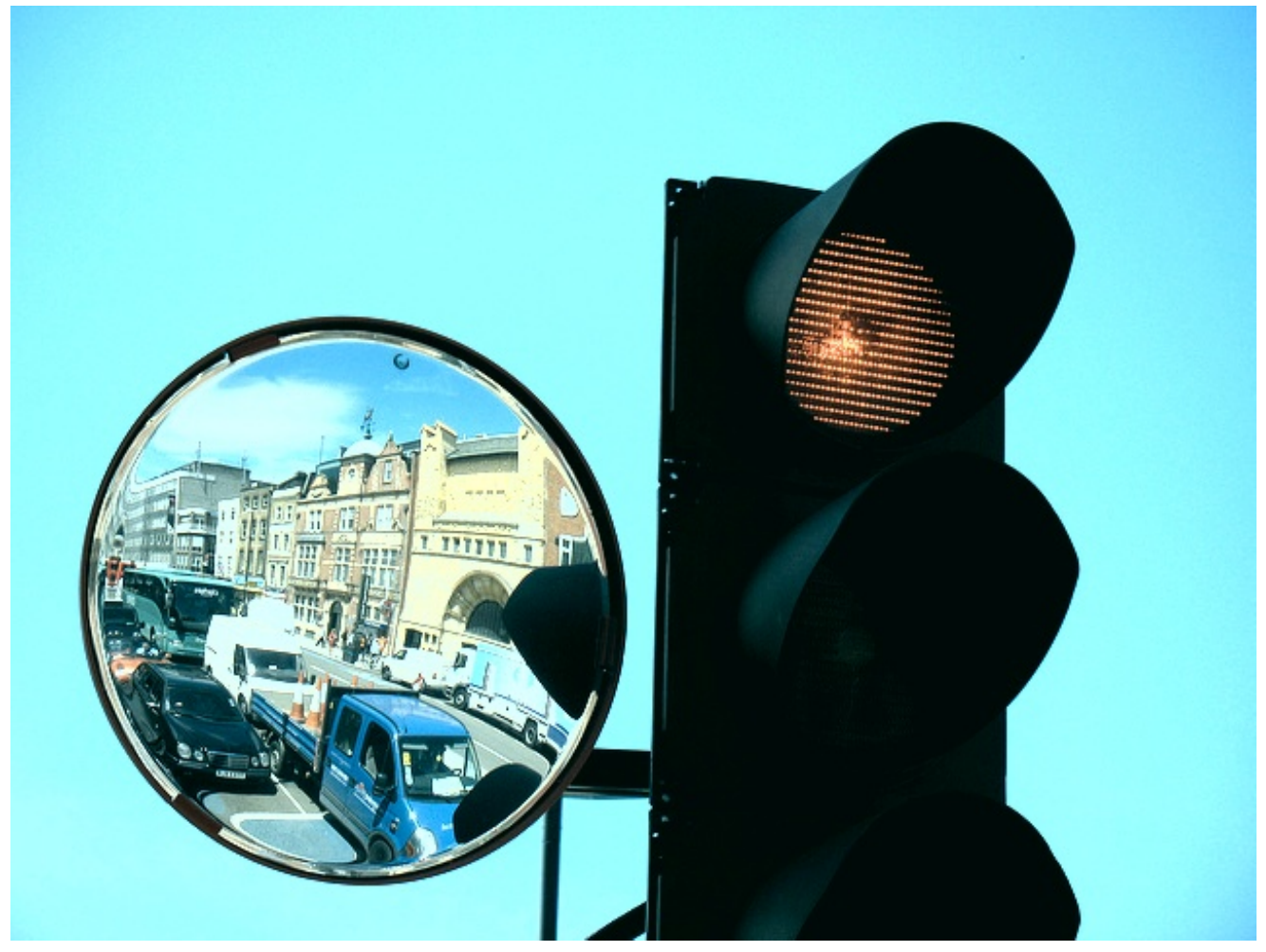}
}
\caption{An example of image recoloring. (a) Natural image. (b) Recolored image generated by the method \cite{reinhard2001color}.}
\label{fig_contrast}
\end{figure}

Different from such common manipulations that change the image content by adding or deleting interest regions, image recoloring aims to change the theme or style of the image by manipulating the color values without compromising the details. Since color information plays a key role in visual understanding, this tampering has the potential to mislead human beings. In addition, for automatic tracking/recognition, recoloring may lead to misjudgment in the artificial vision system, as shown in Fig. \ref{fig_contrast}.


At present, image forensics technologies \cite{verdoliva2020media,farid2009image,stamm2013information} for various types of forgery have appeared in the past decades. According to the operating mechanism, they can be classified into the following two categories: active and passive forensics techniques. Active forensics, e.g., digital watermarking \cite{wang2017hybrid}, embeds identification information into the image in advance. Such embedded data will serve as a marker for integrity verification. However, it requires the operation before the distribution, which limits its practical application. Passive forensics, also called blind forensics, does not rely on such side information, but exclusively on the analyzed multimedia data. It tries to reveal anomalies that may indicate manipulation. For example, source identification utilizes the inherent features of camera hardware and software like lens distortion \cite{yerushalmy2011digital}, color filter array (CFA) \cite{popescu2005exposing}, and pattern noise \cite{cozzolino2019noiseprint}. Specific artifacts left by JPEG compression \cite{lukavs2003estimation}, contrast enhancement \cite{cao2014contrast}, resampling, copy-move \cite{li2018fast}, splicing \cite{matern2019gradient}, and inpainting \cite{li2019localization} are also used for forgery detection. In general, the existing forensics methods have achieved good performance in detecting traditional manipulation. However, due to the different tampering mechanisms, these methods are not satisfactory when they are directly applied to recoloring image detection. As far as we know, there are few forensics works specially designed for the image recoloring, even if altering the color is one of the most common operations in image processing. Yan \textit{et al.} \cite{yan2018recolored} first attempted to take advantage of the consistencies of inter-channel and illumination to distinguish whether an image is recolored. However, the hard-coding operation of the differential image (e.g., $\mathbf{R}-\mathbf{G}$) may not necessarily depict the optimal correlation, and the effectiveness of this method is only verified on the conventional image recoloring approaches. Note that there is currently no work specifically dealing with the deep learning-based recoloring scene.

In this paper, we propose an end-to-end method to distinguish between natural images (NIs) and recolored images (RIs). To our best knowledge, this is the very first work to explore solutions from the perspective of the spatial correlation of adjacent pixels. It exhibits the general ability to detect both conventional and deep learning recoloring forgeries.

Our work started with the idea that recoloring will inevitably destroy the spatial correlation between pixels, due to the inherent difference between the camera imaging and the RI generation. This idea is verified by numerical experiments, where the correlation between their adjacent pixels is statistically discriminative in each color component. Based on such revealed prior knowledge, we design a feature set to identify the RI, using the co-occurrence matrices as the second-order statistics of the spatial correlation between adjacent pixels on each color component. Then, the feature set is fed into a convolution neural network for learning high-level representations, and hence detecting the RI robustly and discriminatively.

To train our proposed network, we introduce three recoloring methods \cite{reinhard2001color,pitie2007automated,yoo2019photorealistic} to generate the training set. We also validate the effectiveness of the proposed method in two challenging detection scenes, i.e., deep learning-based recoloring scene and conventional recoloring scene. In the deep learning-based recoloring scene, our trained network achieves satisfactory generalization on the testing set generated by the other four deep learning-based recoloring methods \cite{lee2020deep,luan2017deep,afifi2019image,li2018closed}. In the conventional recoloring scene, our trained network also exhibits favorable performance on the testing set released by Yan \textit{et al.} \cite{yan2018recolored}, indicating that proposed method is also suitable for conventional RIs.

The main contributions of this paper are summarized as follows:
\begin{itemize}
\item We are the first to design a recoloring detection method by spatial correlation, which shows the general ability to detect both conventional and deep learning recoloring.
\item We theoretically analyze that the recoloring operation will inevitably destroy the spatial correlation between pixels of NIs. Moreover, numerical experiments verify the possibility of spatial correlation as a discriminative feature. 
\item We propose an effective feature set for recoloring detection, which formed on the co-occurrence matrices extracted from each color component of the image.
\item We generate a large-scale and high-quality benchmark with various recoloring methods, which can support the future research on recoloring detection. 
\end{itemize}

The rest of this paper is outlined below. Section \ref{section2} introduces related work. The details of our method are presented in Section \ref{section3}. In Section \ref{section4}, we show the experimental results in various scenes. The last section concludes this work and highlights some promising directions.

\section{Related Work}
\label{section2}
In this section, recoloring approaches and related detection methods are reviewed accordingly.

\subsection{Recoloring approaches}
 So far, a variety of image recoloring approaches have been developed. According to the operating mechanism, most of the image recoloring techniques can be categorized into four classes, example-based (color transfer) \cite{reinhard2001color,pitie2005n,he2019progressive,lee2020deep,tai2005local}, stroke-based (edit propagation) \cite{levin2004colorization,qu2006manga,an2008appprop,xu2009efficient,chen2014sparse,endo2016deepprop,zhang2017real}, palette-based \cite{o2011color,lin2013probabilistic,chang2015palette,cho2017palettenet,afifi2019image}, and photorealistic style transfer \cite{luan2017deep,li2018closed,yoo2019photorealistic}. These techniques are briefly summarized in Table \ref{tab:summary recoloring approach}.

\begin{table*}[htbp]
\centering
\scriptsize
\caption{Summary of the Existing Recoloring Approaches}
\label{tab:summary recoloring approach}
\begin{tabular}{m{4.5cm}<{\centering} m{6cm}<{\centering} m{1.25cm}<{\centering} m{2cm}<{\centering}} 
\toprule
\textbf{Category} & \textbf{Definition} & \textbf{Method} & \textbf{Side information}  \\ \midrule
    Example-based recoloring (color transfer) & The exampled-based recoloring approaches aim to transfer the color style of the reference image to the source image &  \cite{reinhard2001color,pitie2005n,he2019progressive,lee2020deep,tai2005local} & Reference image \\
    Stroke-based recoloring (edit propagation) & The stroke-based recoloring approaches first draw scribbles with the desired color in different regions, and then automatically propagate these edits to similar pixels &  \cite{levin2004colorization,qu2006manga,an2008appprop,xu2009efficient,chen2014sparse,endo2016deepprop,zhang2017real} & User scribbles \\
    Palette-based recoloring & The palette-based approaches generate a suitable palette for an input image, and then users can manipulate the image by modifying the colors in the palette & \cite{o2011color,lin2013probabilistic,chang2015palette,cho2017palettenet,afifi2019image} &  Select the palette manually \\
    Photorealistic style transfer recoloring & The photorealistic style transfer approaches can transfer the style in a photorealistic way &  \cite{luan2017deep,li2018closed,yoo2019photorealistic} & Reference image \\ \bottomrule
\end{tabular}
\end{table*}

\subsubsection{Example-based recoloring (color transfer)}
The exampled-based recoloring approaches aim to transfer the color style of the reference image to the source image. Early methods \cite{reinhard2001color,pitie2005n} apply the characteristics of the reference image to the source image by analyzing the global color distribution, which belongs to global color transfer. In addition, the local color transfer method \cite{tai2005local} first identifies the regional correlation between the source image and the reference image, then transfers the color distribution between the corresponding regions. Different from conventional color transfer methods described above that match low-level features, deep learning-based approaches \cite{he2019progressive,lee2020deep} can reflect higher-level semantic relationships.

\subsubsection{Stroke-based recoloring (edit propagation)}
The stroke-based recoloring approaches first draw scribbles with the desired color in different regions, and then automatically propagate these edits to similar pixels. In most of the conventional stroke-based works \cite{levin2004colorization,qu2006manga,an2008appprop,xu2009efficient,chen2014sparse}, users must heuristically determine the image feature they use and adjust parameters for the features based on their own needs and target images. In contrast, deep learning-based approaches \cite{} can automatically learn appropriate feature representation from low-level clues, high-level semantic information, and user strokes.

\subsubsection{Palette-based recoloring}
The palette-based approaches generate a suitable palette for an input image, and then users can manipulate the image by modifying the colors in the palette. Most of the conventional works \cite{o2011color,lin2013probabilistic,chang2015palette} create palettes by the \textit{k}-means clustering algorithm or its variants. However, the color transformation function in traditional palette space cannot be used for content-aware recoloring. Therefore, deep learning-based methods \cite{cho2017palettenet,afifi2019image} take advantage of neural networks in understanding the content of the source image and embed the palette into the object segmentation network \cite{ronneberger2015u}.

\subsubsection{Photorealistic style transfer recoloring}
The photorealistic style transfer approaches \cite{luan2017deep,li2018closed,yoo2019photorealistic} are derived from the neural style transfer \cite{Gatys_2016_CVPR,li2017universal} in recent years. Unlike the style conversion methods, the output still looks like an art painting, even though both the input and reference style images are photos. The photorealistic style transfer approaches utilize the whitening and coloring transforms to prevent spatial distortion and limit transmission operations to occur only in color space, resulting in a satisfying photorealistic style rather than being like a painting.

\subsection{Recoloring detection method}
Very recently, Yan \textit{et al.} \cite{yan2018recolored} made the first effort to detect image recoloring. They observed that the inter-channel correlation and illumination consistency may not be maintained after the image is recolored. In detail, this work designed a deep discriminative model based on the VGG network \cite{simonyan2014very}, which is composed of three feature extractors and a feature fusion module. The original RGB image with its differential image (DI) and illumination image (IM) are utilized as three inputs of the network. After extracting the forgery-related features, a feature fusion module is employed to refine these features and output the forgery probability. For training the network, they used the conventional recoloring methods \cite{reinhard2001color,beigpour2011object,pitie2007automated} to generate a large-scale training set. The testing results on a new synthetic dataset (generated by a variety of conventional recoloring methods \cite{reinhard2001color,beigpour2011object,pitie2007automated,chang2015palette,pitie2007linear,grogan2015l2,an2008appprop}) and a newly collected dataset (downloaded from websites or made by mobile applications) also demonstrate the effectiveness of their method. However, the hard-coded feature of differential image (e.g., $\mathbf{R}-\mathbf{G}$) may not be the best feature to describe the inter-channel correlation. From their experimental results, this method may not be able to cope with the recoloring scene of deep learning method. In addition, to the best of our knowledge, no method has yet been developed to detect the RIs generated by deep learning-based recoloring methods \cite{lee2020deep,he2019progressive,endo2016deepprop,zhang2017real,cho2017palettenet,afifi2019image,luan2017deep,li2018closed,yoo2019photorealistic}. 


\begin{figure*}[htbp]
\centering
\subfigure[Natural image acquisition in digital camera]{
\includegraphics[scale=0.58]{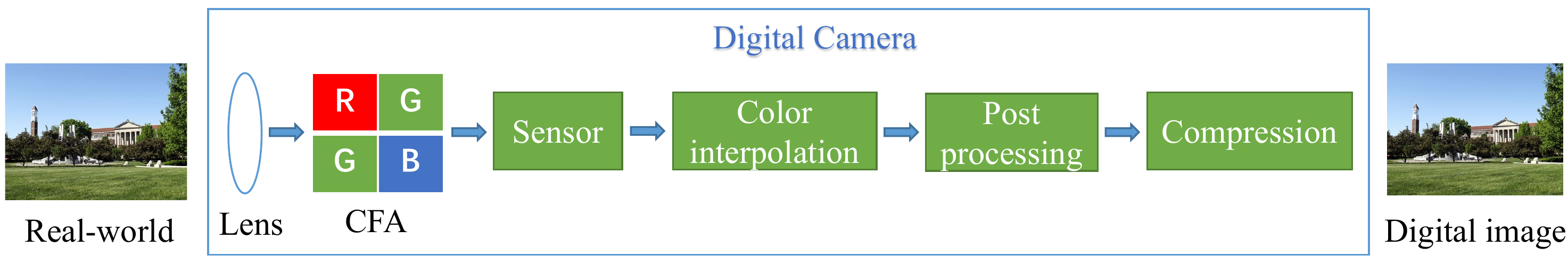}
}
\quad
\subfigure[Recoloring image generation by photorealistic style transfer]{
\includegraphics[scale=0.6]{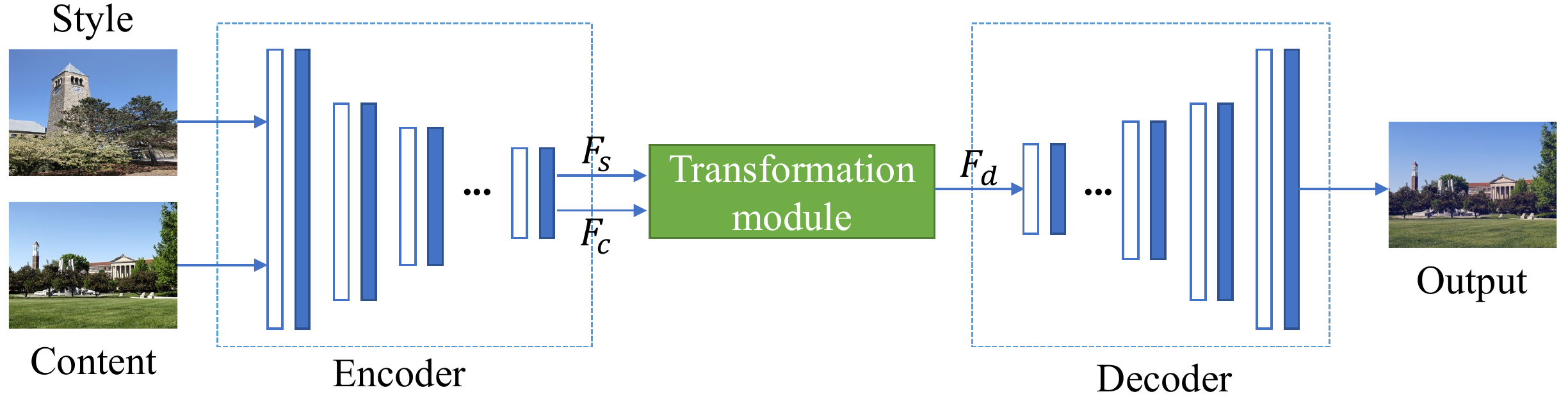}
}
\caption{The illustration of natural image acquisition and recoloring image generation.}
\label{fig_imaging}
\end{figure*}

\begin{figure*}[htbp]
\centering
\includegraphics[scale=0.55]{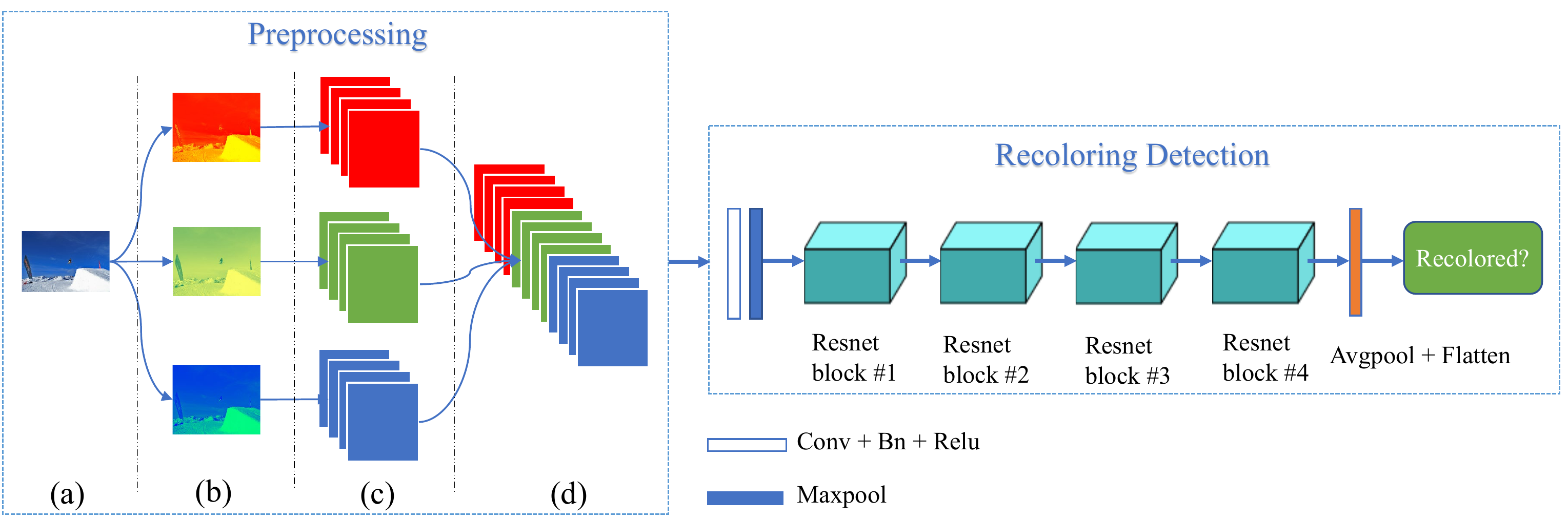}
\caption{The pipeline of the proposed method. In preprocessing module, the input image (a) is split into its three color channels (b). For each color channel, co-occurrence matrices in four directions are calculated. These matrices (c) are then stacked to produce a single tensor (d) for recoloring detection. The CNN network in recoloring detection module is adapted from Resnet18 \cite{he2016deep}.}
\label{fig_framework}
\end{figure*}

\section{Methodology}
\label{section3}
\subsection{Motivation}
\subsubsection{The difference between camera imaging and RI generation}
To distinguish RIs from NIs, we first need to understand how NI acquisition and RI generation. As indicated in Fig. \ref{fig_imaging} (a), the general process of acquiring NIs by a digital camera is as follows. The incoming light first passes through optical filters to retain the desired light components and then is focused on the charge-coupled device or complementary metal-oxide-semiconductor sensor by the lens. Under the action of CFA on the sensor, only one color (red, green, or blue) is obtained for each pixel, and other colors must be obtained by CFA interpolation based on the color information of surrounding pixels. Finally, a digital image is obtained through a series of signal processing operations such as white balance, color processing, image sharpening, contrast enhancement, gamma correction, and compression.

By contrast, the way of generating images by various recoloring methods is very different. For example, the photorealistic style transfer can be regarded as an image reconstruction problem with feature projection, the general process is shown in Fig. \ref{fig_imaging} (b). Its framework consists of two parts: encoder and decoder. First, the encoder extracts the vectorized features of the image through a series of convolution and pooling operations. Then, the vectorized features $F_ {s} $, $F_ {c} $ of the style image and the content image are respectively input to the conversion module, and the converted vectorized features $F_ {d}$ are outputted. Finally, the decoder decodes $F_{d}$ to obtain an output image. The CFA interpolation makes the pixel value of NIs have a special correlation, while the generation of RIs is different from the capture of NIs, and there are no color processing operations similar to CFA. Therefore, the image recoloring operation will inevitably destroy this correlation. This hypothesis will be supported with some experimental evidence in the following analysis. 

\subsubsection{Discriminability of spatial correlation}
\label{section3.1.2}
While the visual results of the RI are reasonable, the recoloring operation may alter the statistical features of NI pixels, such as the spatial correlation between pixels. In the three color channels (R, G, B) of RIs and NIs, the correlation between adjacent pixels will be considered for analysis. A measure will be utilized to show the statistical difference between RIs and NIs.

For the \textit{i}-th image $\mathbf{I}$ in a dataset, we calculate the correlation coefficient between two adjacent pixels in the four directions (horizontal, vertical, diagonal and anti-diagonal) in each color component $\mathbf{I}^c$ ($c \in \{\mathbf{R}, \mathbf{G}, \mathbf{B}\}$). For the sake of brevity, we only take vertical correlation as an example here, and other directions are similar. The correlation formula is:
\begin{equation}
\begin{small}
\label{correlation-coefficient}
r_{i}^{c}=\frac{\sum_{j=1}^{w} \sum_{k=1}^{h-1}\left(\mathbf{I}_{j, k}^{c}-\overline{\mathbf{I}}^{c}\right)\left(\mathbf{I}_{j, k+1}^{c}-\overline{\mathbf{I}}^{c}\right)}
{\sqrt{\sum_{j=1}^{w} \sum_{k=1}^{h-1}\!\left(\mathbf{I}_{j, k}^{c}\!-\overline{\mathbf{I}}^{c}\right)^{2}\! \sum_{j=1}^{w}\! \sum_{k=1}^{h-1}\!\left(\mathbf{I}_{j, k+1}^{c}\!-\overline{\mathbf{I}}^{c}\!\right)^{2}}},
\end{small}
\end{equation}
where \textit{w} and \textit{h} are the width and height of the image, $\overline{\mathbf{I}}^{c}$ is the mean value of $\mathbf{I}^c$, and $r_{i}^{c}$ represents the correlation coefficient between adjacent pixels of the \textit{i}-th image in color component \textit{c}. Specifically, a higher $r_{i}^{c}$ means that adjacent pixels in $\mathbf{I}^c$ have a stronger correlation.

For a recoloring dataset, we first calculate the correlation coefficient $r_{i}^{c}$ of each image $\mathbf{I}$ in each color component $c$. Then a histogram is constructed based on this coefficient, denoted as $\mathbb{H}_{\mathrm{rec}}^{c}$. Similarly, for a set of NIs, we construct the histogram $\mathbb{H}_{\mathrm{nat}}^{c}$. Then, we use the chi-square distance to measure the similarity between the histograms $\mathbb{H}_{\mathrm{rec}}^{c}$ and $\mathbb{H}_{\mathrm{nat}}^{c}$:
\begin{equation}
\label{chi-square}
d_{\chi^{2}}\left(\mathbb{H}_{\mathrm{rec}}^{c}, \mathbb{H}_{\text {nat}}^{c}\right)=\frac{1}{2} \sum_{x} \frac{\left(\mathbb{H}_{\mathrm{rec}}^{c}(x)-\mathbb{H}_{\mathrm{nat}}^{c}(x)\right)^{2}}{\mathbb{H}_{\mathrm{rec}}^{c}(x)+\mathbb{H}_{\mathrm{nat}}^{c}(x)},
\end{equation}
where \textit{x} is the bin index of histogram and $d_{\chi^{2}}\left(\mathbb{H}_{\mathrm{rec}}^{c}, \mathbb{H}_{\text {nat}}^{c}\right)$ can serve as a \textit{discriminative metric} reflecting the statistical difference between RIs and NIs. Specifically, a higher $d_{\chi^{2}}\left(\mathbb{H}_{\mathrm{rec}}^{c}, \mathbb{H}_{\text {nat}}^{c}\right)$ means that there is a stronger statistical differences between RIs and NIs.

\begin{figure*}[htbp]
\centering
\subfigure[$d_{\chi^{2}}=0.0205$]{
\includegraphics[width=3.5cm]{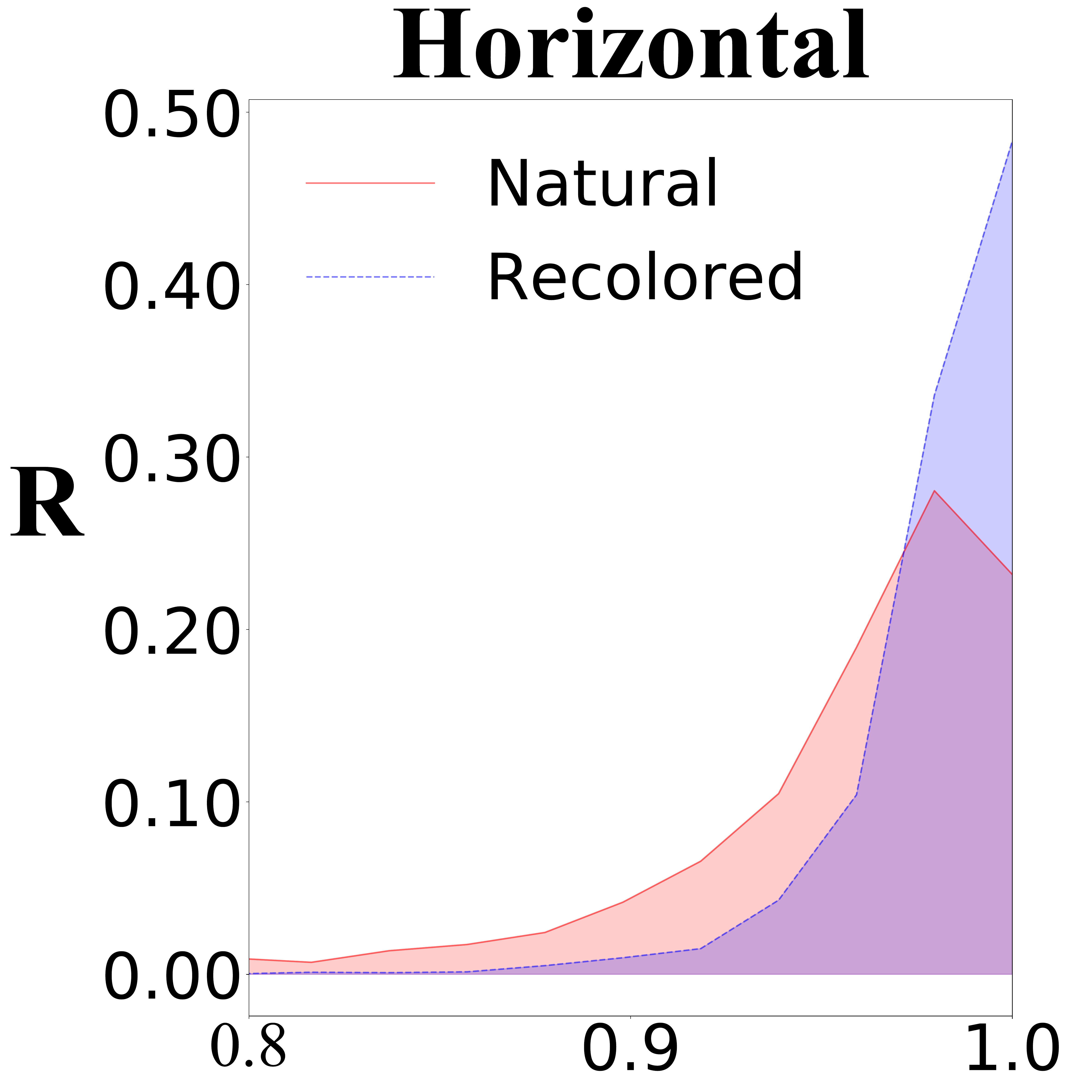}
}
\subfigure[$d_{\chi^{2}}=0.0095$]{
\includegraphics[width=3.5cm]{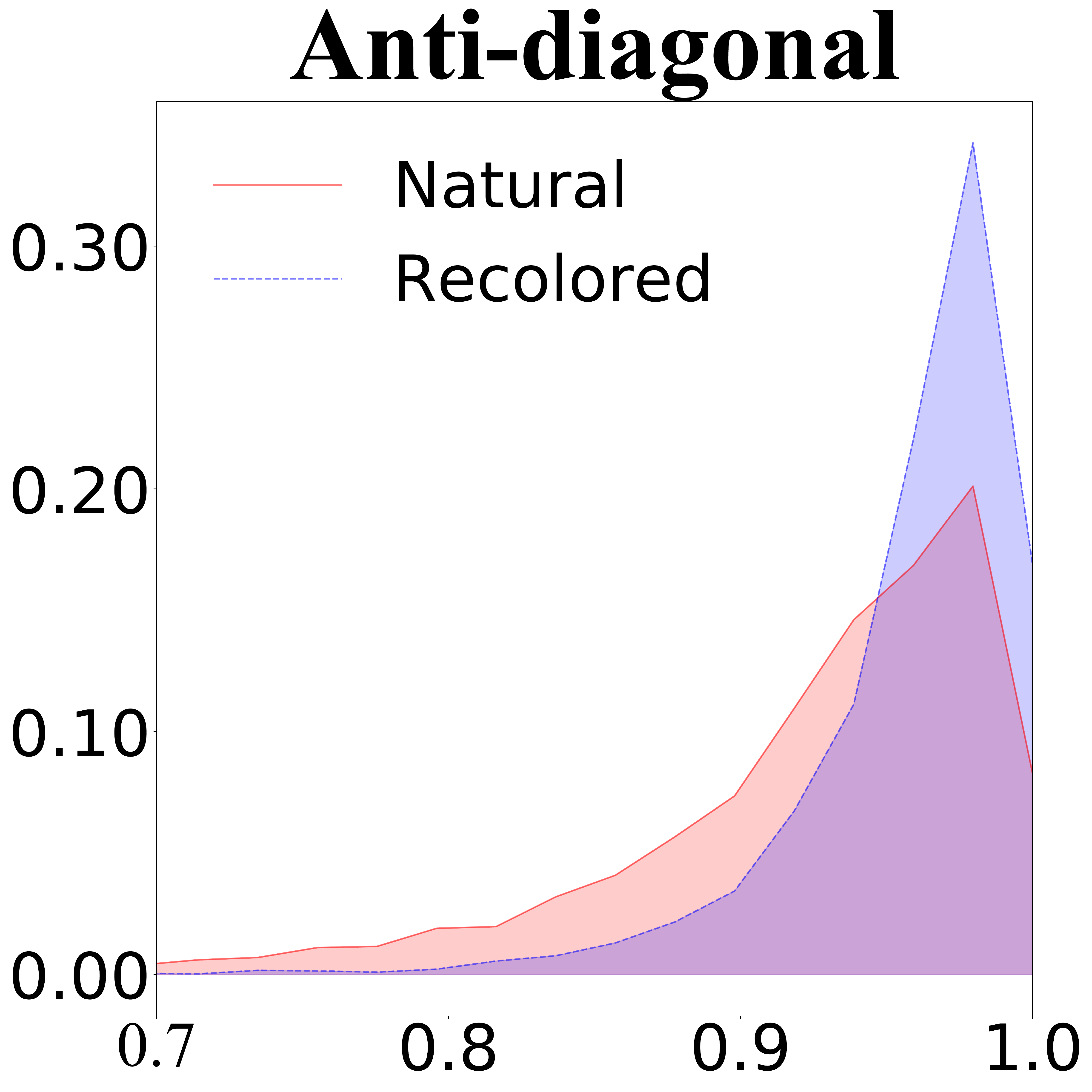}
}
\subfigure[$d_{\chi^{2}}=0.0176$]{
\includegraphics[width=3.5cm]{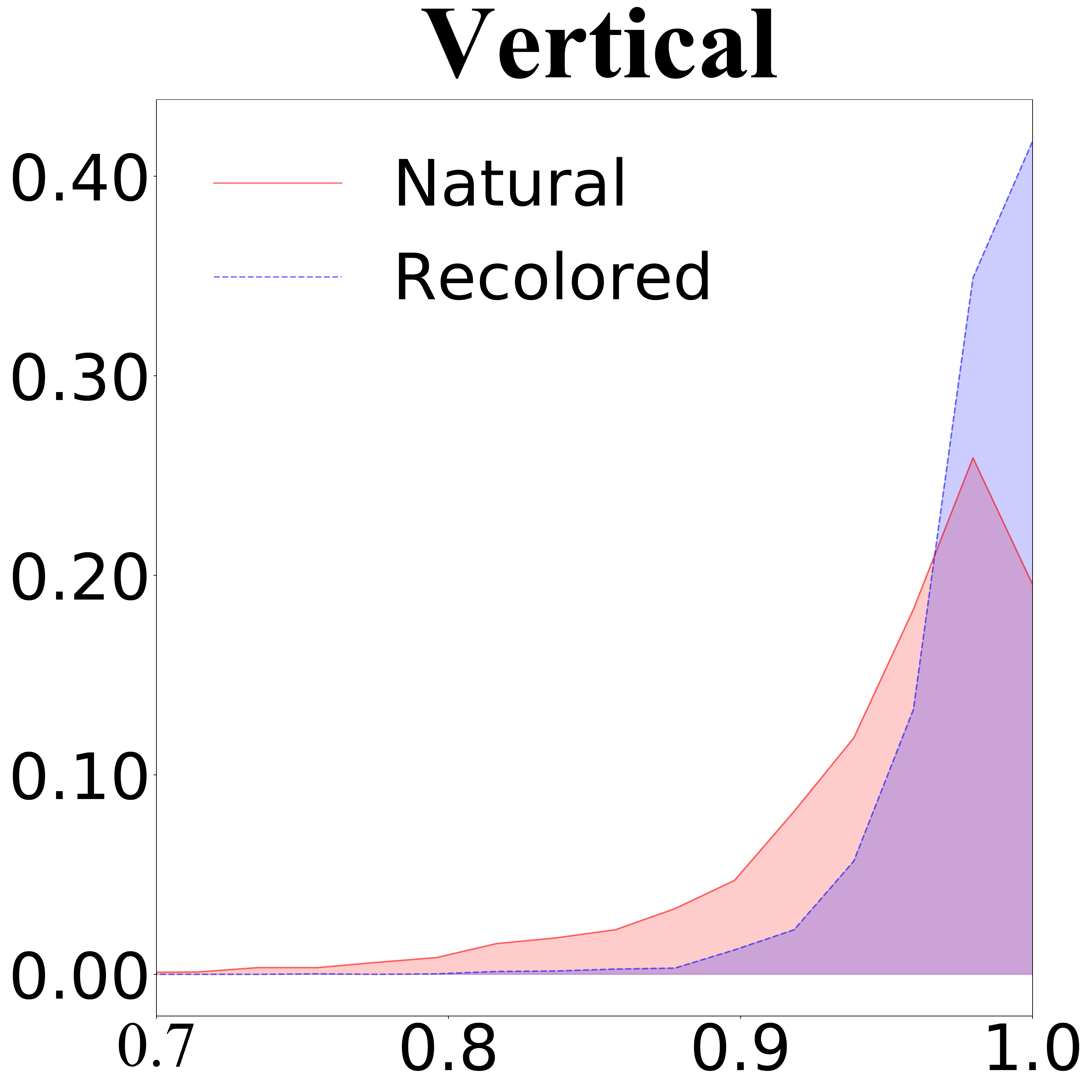}
}
\subfigure[$d_{\chi^{2}}=0.0098$]{
\includegraphics[width=3.5cm]{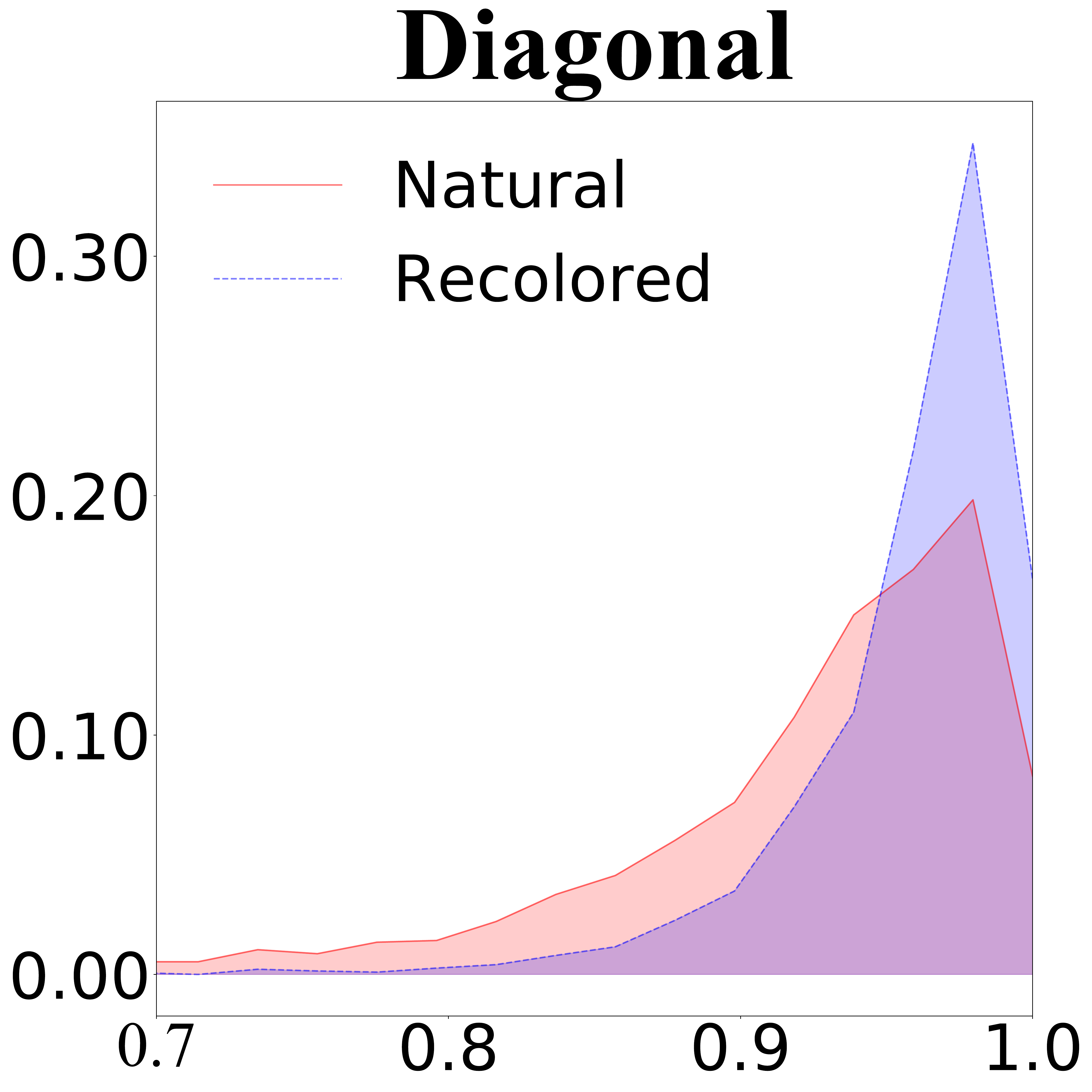}
}
\subfigure[$d_{\chi^{2}}=0.0140$]{
\includegraphics[width=3.5cm]{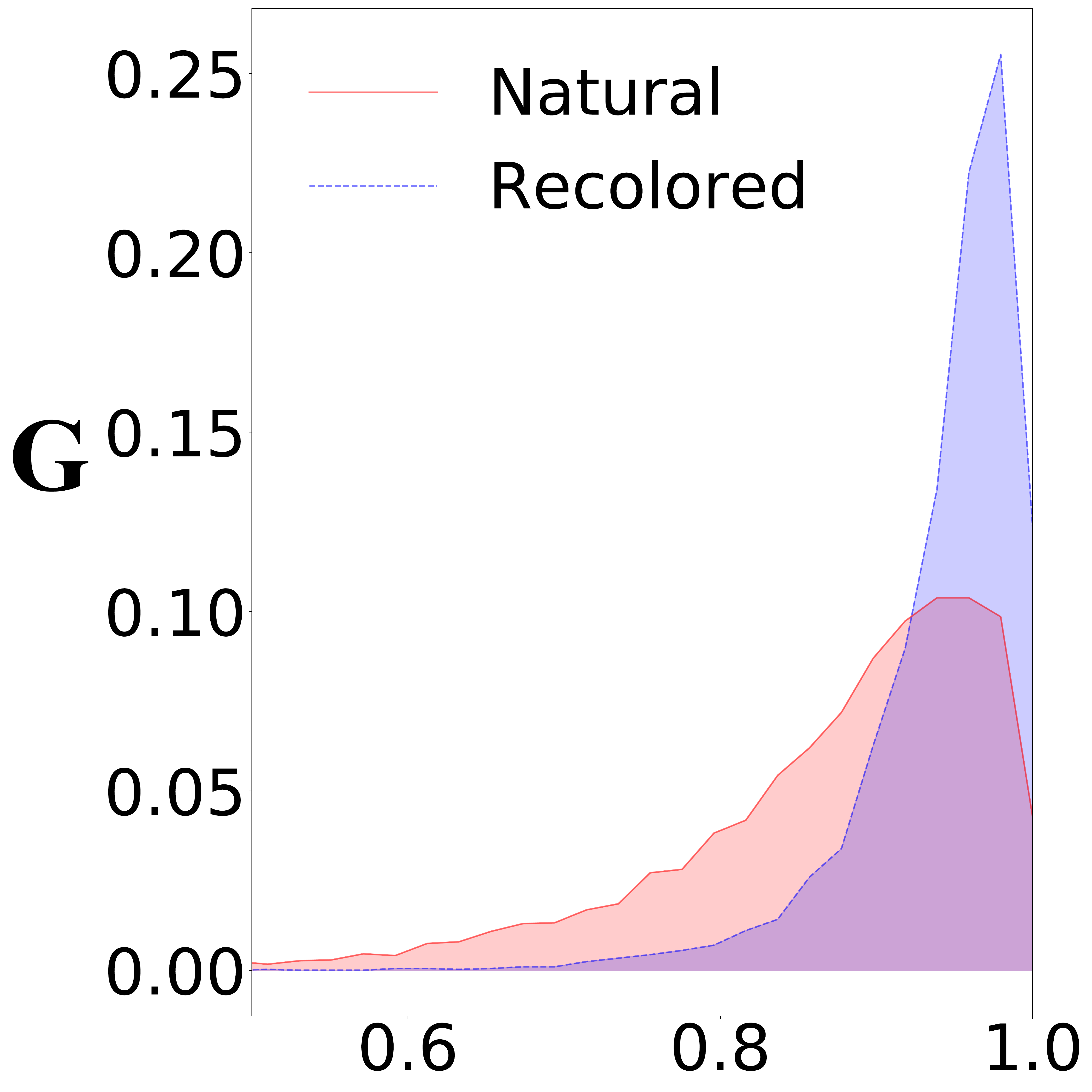}
}
\subfigure[$d_{\chi^{2}}=0.0081$]{
\includegraphics[width=3.5cm]{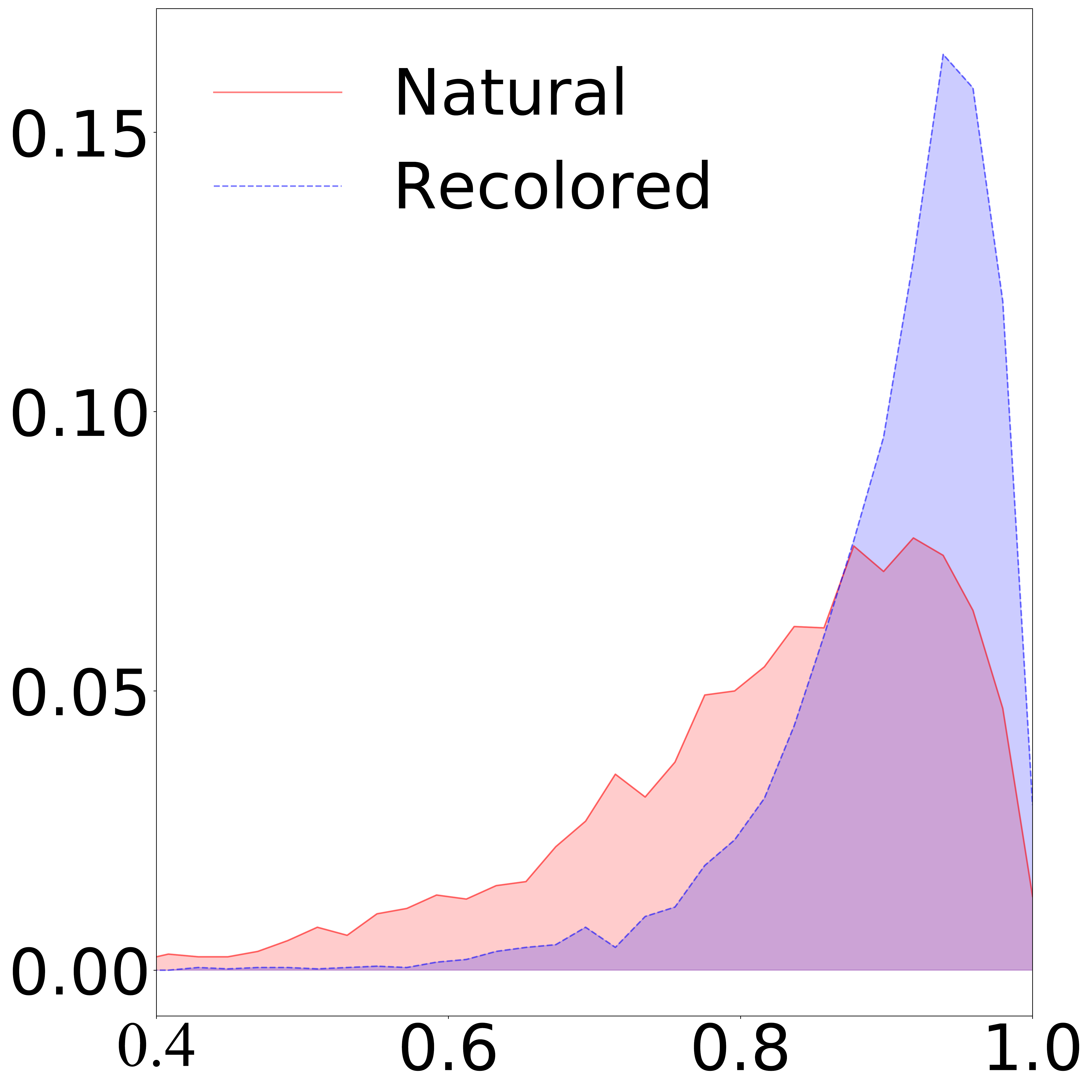}
}
\subfigure[$d_{\chi^{2}}=0.0130$]{
\includegraphics[width=3.5cm]{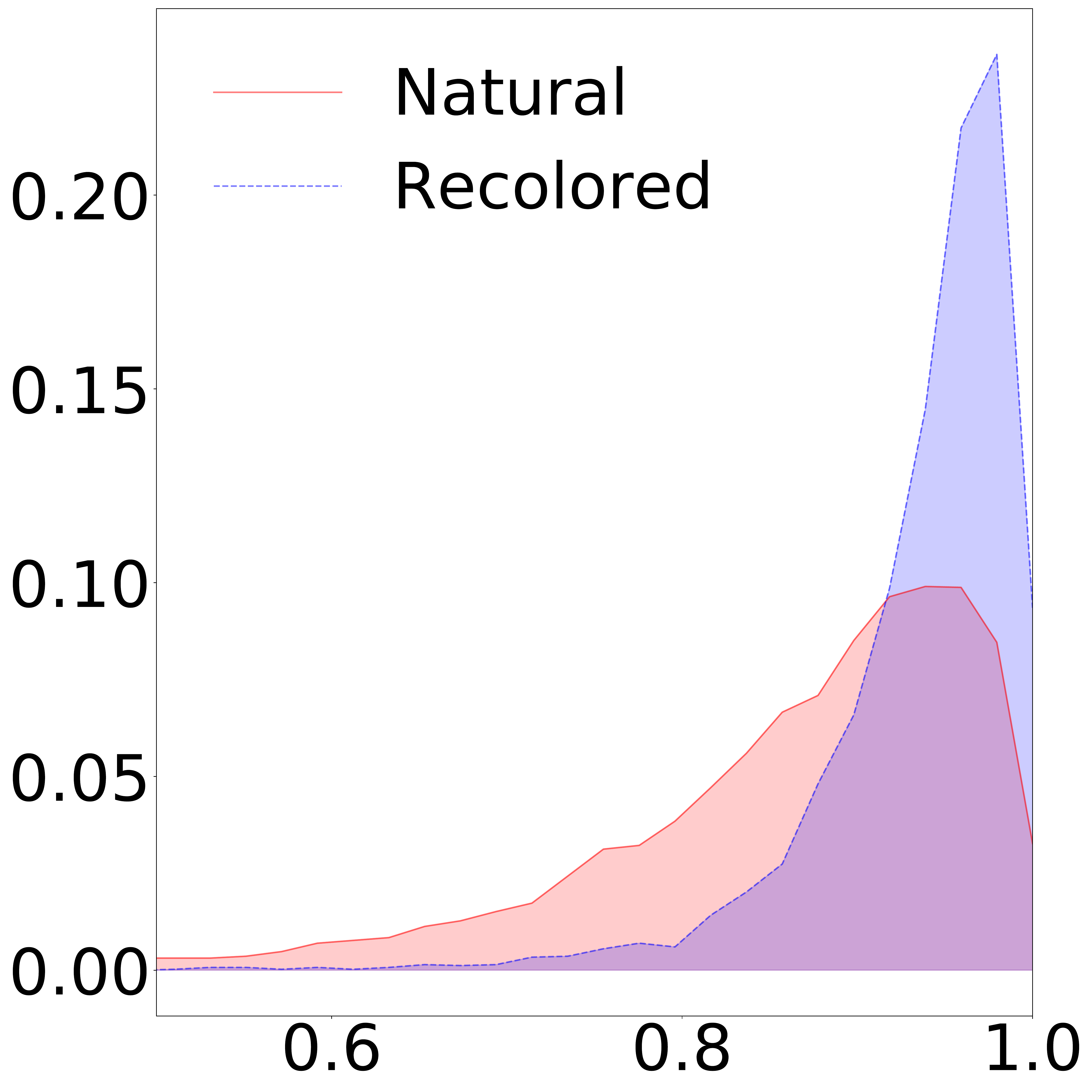}
}
\subfigure[$d_{\chi^{2}}=0.0079$]{
\includegraphics[width=3.5cm]{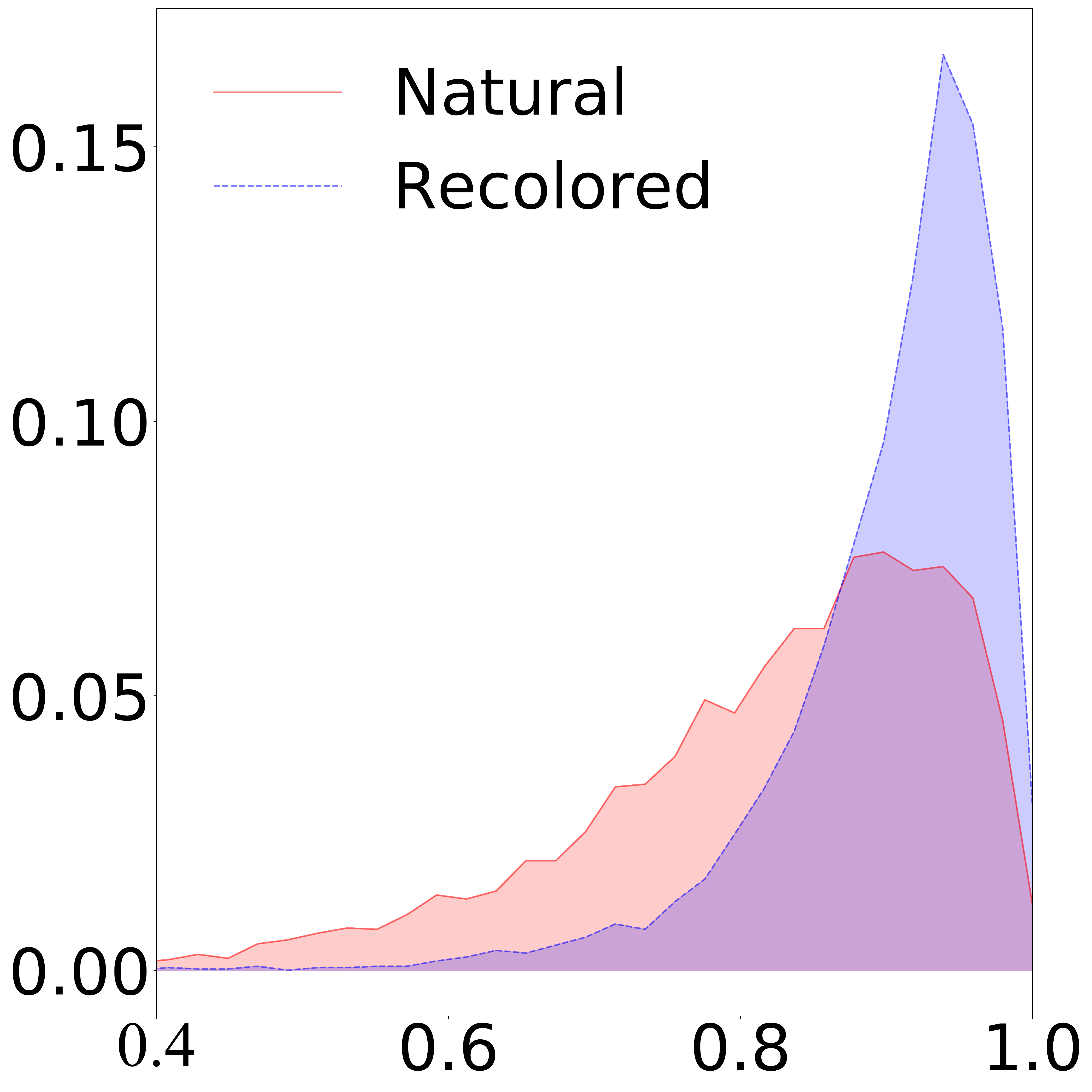}
}
\subfigure[$d_{\chi^{2}}=0.0052$]{
\includegraphics[width=3.5cm]{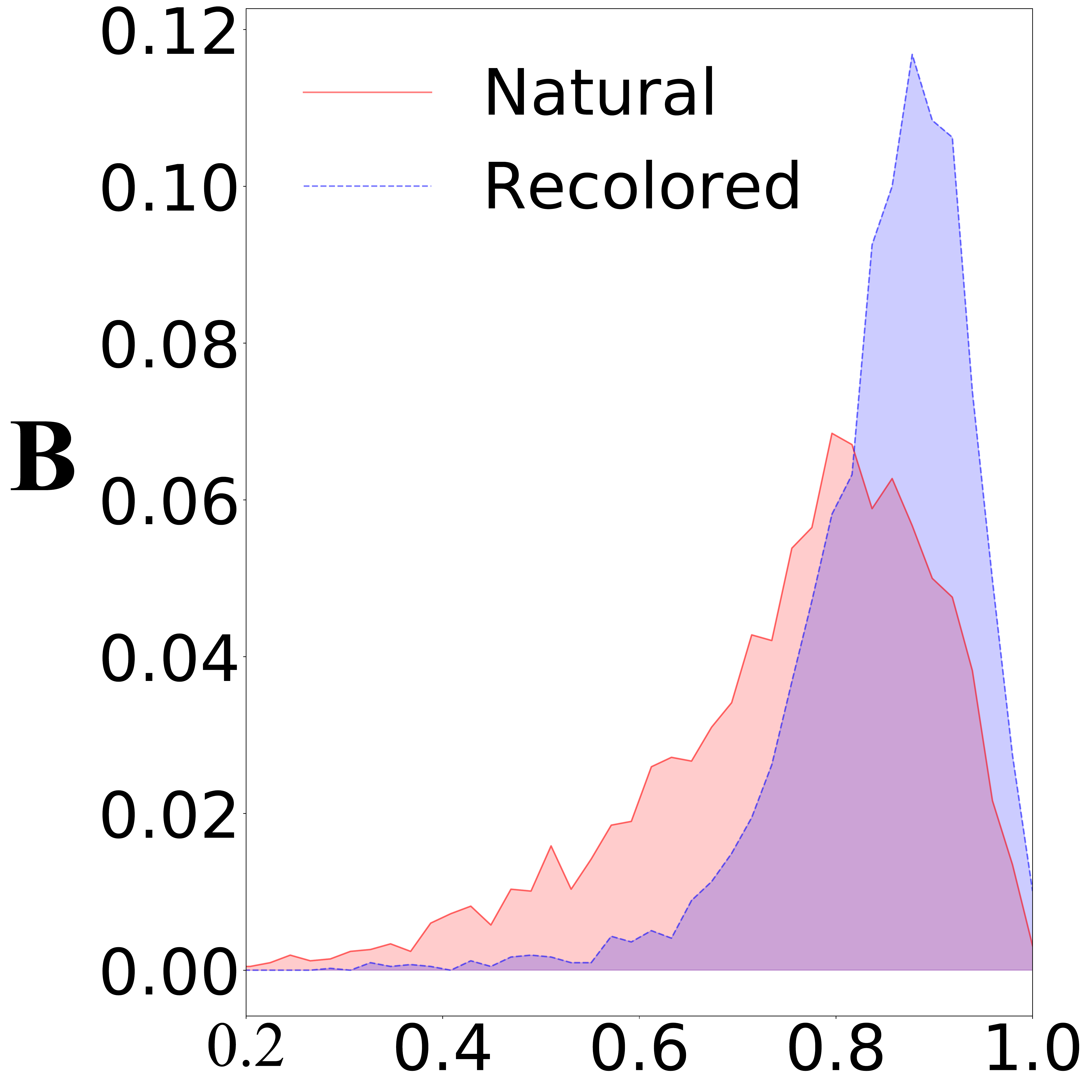}
}
\subfigure[$d_{\chi^{2}}=0.0043$]{
\includegraphics[width=3.5cm]{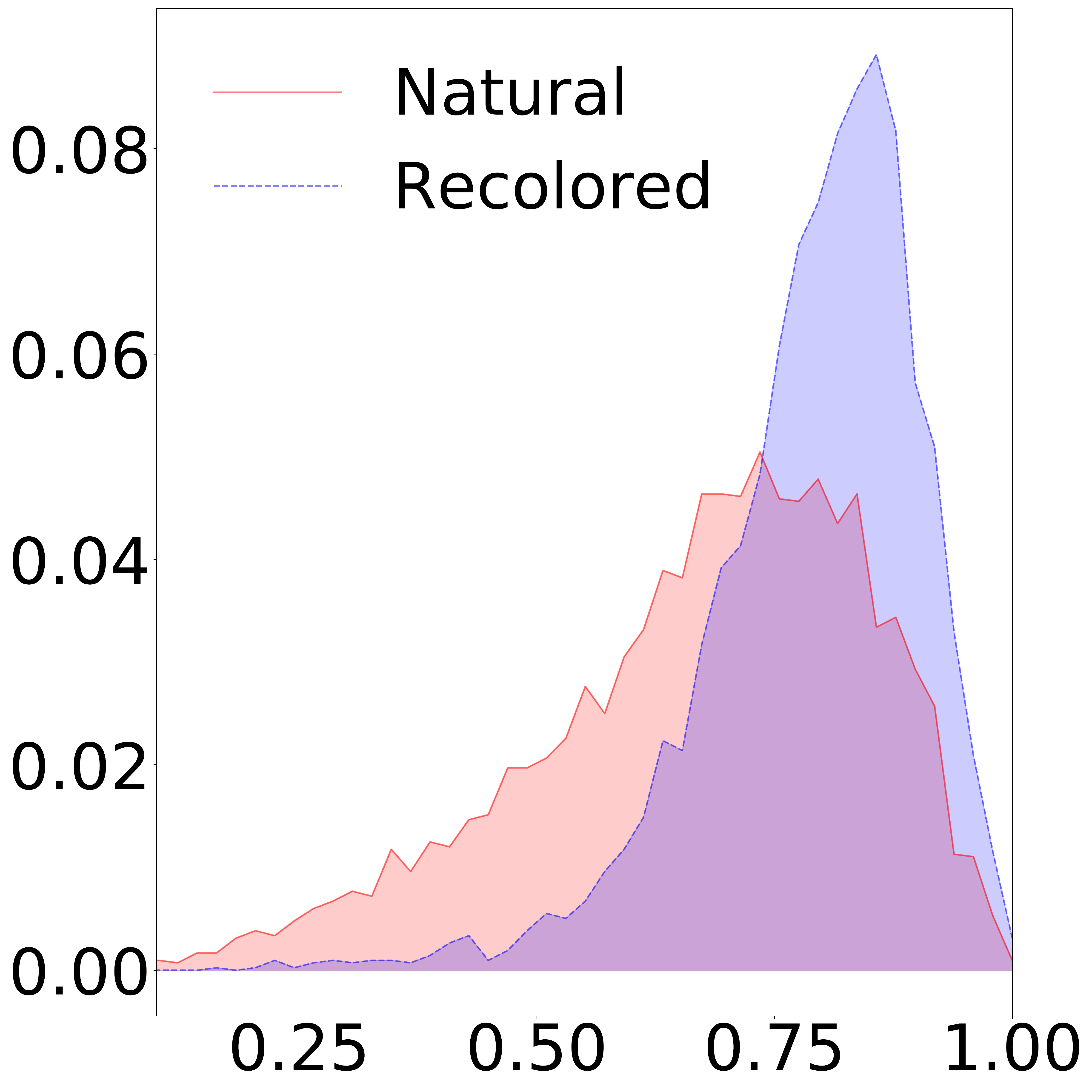}
}
\subfigure[$d_{\chi^{2}}=0.0055$]{
\includegraphics[width=3.5cm]{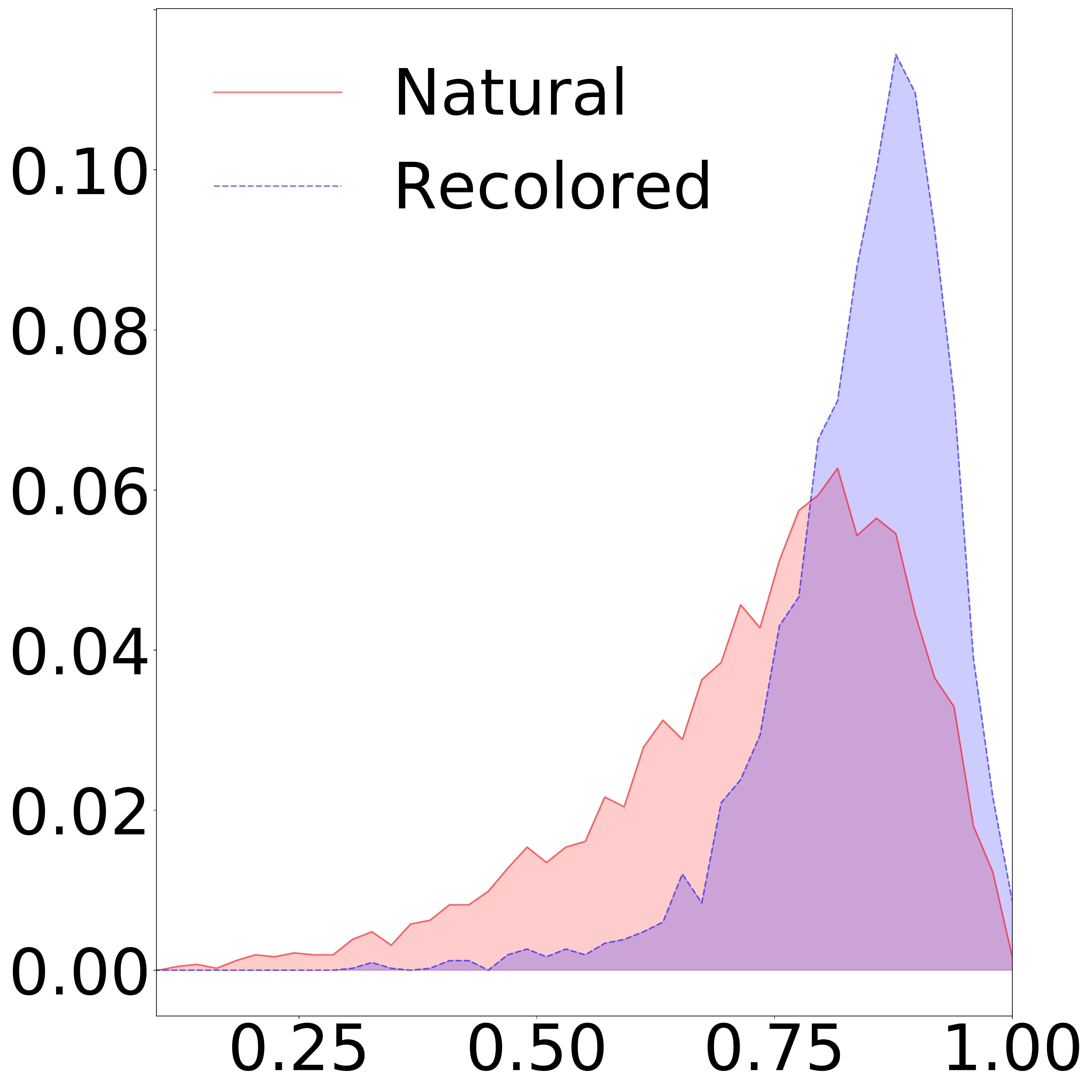}
}
\subfigure[$d_{\chi^{2}}=0.0042$]{
\includegraphics[width=3.5cm]{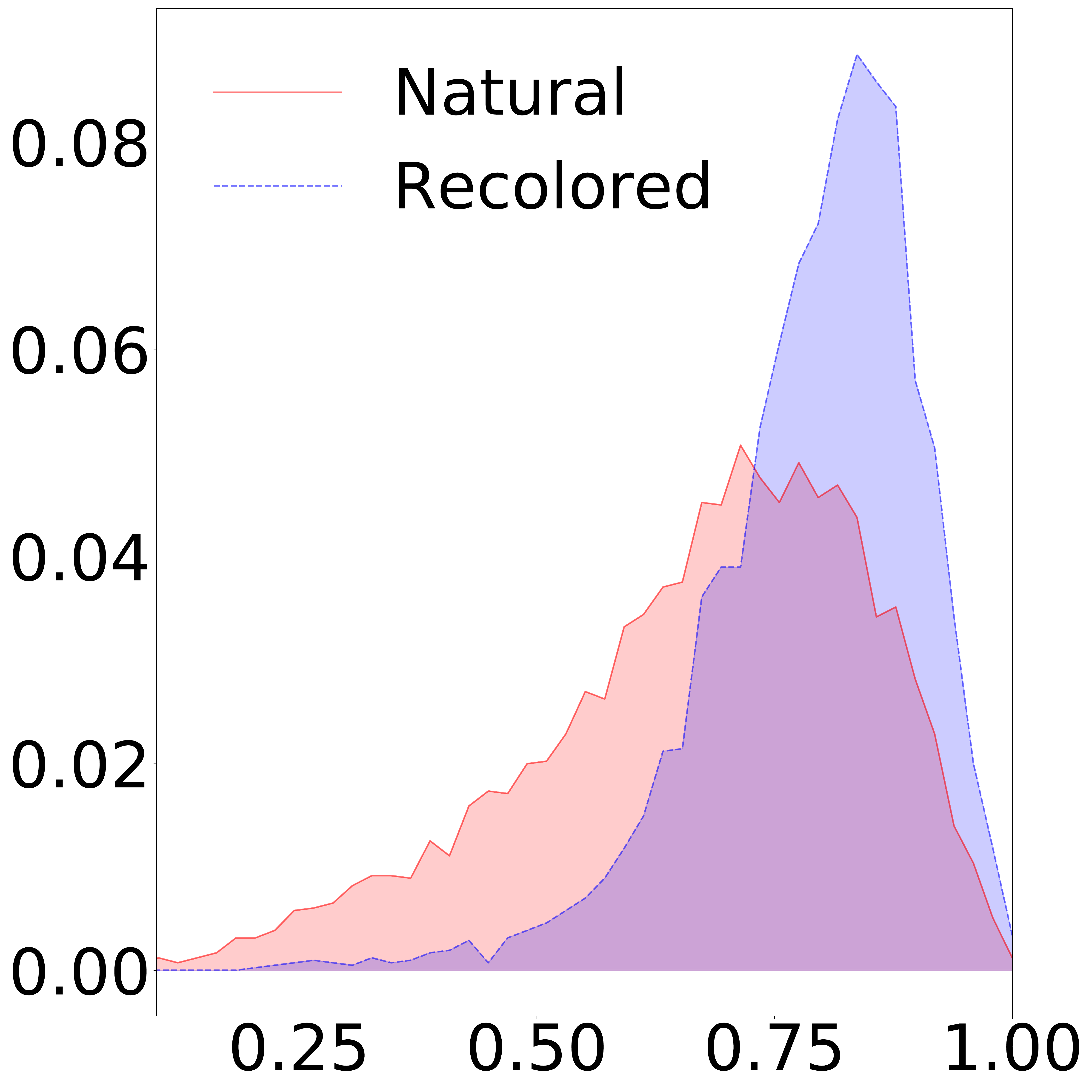}
}
\caption{ The histograms $\mathbb{H}_{\mathrm{rec}}^{c}$ (blue) and $\mathbb{H}_{\mathrm{nat}}^{c}$ (red) in different direction and different color channel. Here, the abscissa of the histogram is $r_{i}^{c}$ bin in the image, and the ordinate represents the frequency of appearance of each bin. Columns 1--4 in the figure are horizontal, anti-diagonal, vertical, and diagonal directions, respectively. Rows 1--3 are the R, G, and B color channels, respectively. The values of $d_{\chi^{2}}$ are included in the sub-captions.}
\label{chi-square-figure}
\end{figure*}

Next, we will conduct several experiments to evaluate the discriminative metric. The COCO validation dataset \cite{lin2014microsoft} is utilized as NIs. Approximately 16,000 RIs are generated by the method \cite{yoo2019photorealistic}. Then 5,000 RIs and their corresponding NIs are randomly selected for analysis. First, we calculate $r_{i}^{c}$ ($c \in \{\mathbf{R}, \mathbf{G}, \mathbf{B}\}$) in the four directions (horizontal, vertical, diagonal, and anti-diagonal) as shown in (\ref{correlation-coefficient}). Then, we apply (\ref{chi-square}) to construct the histograms $\mathbb{H}_{\mathrm{rec}}^{c}$ and $\mathbb{H}_{\text {nat}}^{c}$. The histograms for different color channels and different directions are shown in Fig. \ref{chi-square-figure}. The non-overlapping regions of $\mathbb{H}_{\mathrm{rec}}^{c}$ and $\mathbb{H}_{\text {nat}}^{c}$ reflect the difference between the RIs and the NIs. In the sub-captions of the corresponding sub-figures in Fig. \ref{chi-square-figure}, we observe that the values of $d_{\chi^{2}}\left(\mathbb{H}_{\mathrm{rec}}^{c}, \mathbb{H}_{\text {nat}}^{c}\right)$ for different color channels and different directions are all exceeded 0.004. In particular, the horizontal and vertical $d_{\chi^{2}}\left(\mathbb{H}_{\mathrm{rec}}^{c}, \mathbb{H}_{\text {nat}}^{c}\right)$ values in R and G color channels are both greater than 0.01. These results indicate a statistical difference between NIs and RIs in terms of spatial correlation, leading to potential uses in detecting recolored image.

\subsection{Method Details}
\subsubsection{Overview}
We propose an end-to-end pipeline to detect the recolored image, as illustrated in Fig. \ref{fig_framework}. Firstly, the co-occurrence matrices are calculated for each color channel of the image in the preprocessing module. Such matrices represent the spatial correlation characteristics between adjacent pixels. Then, unlike the previous method, a machine learning classifier is adopted to classify different images. We feed the tensor composed of the co-occurrence matrices through the CNN to automatically learn important features and reveal image recoloring operations. We elaborate on the technical details in the following.

\subsubsection{Extracting spatial correlation features}
\begin{figure}[htbp]
\centering
\includegraphics[scale=0.4]{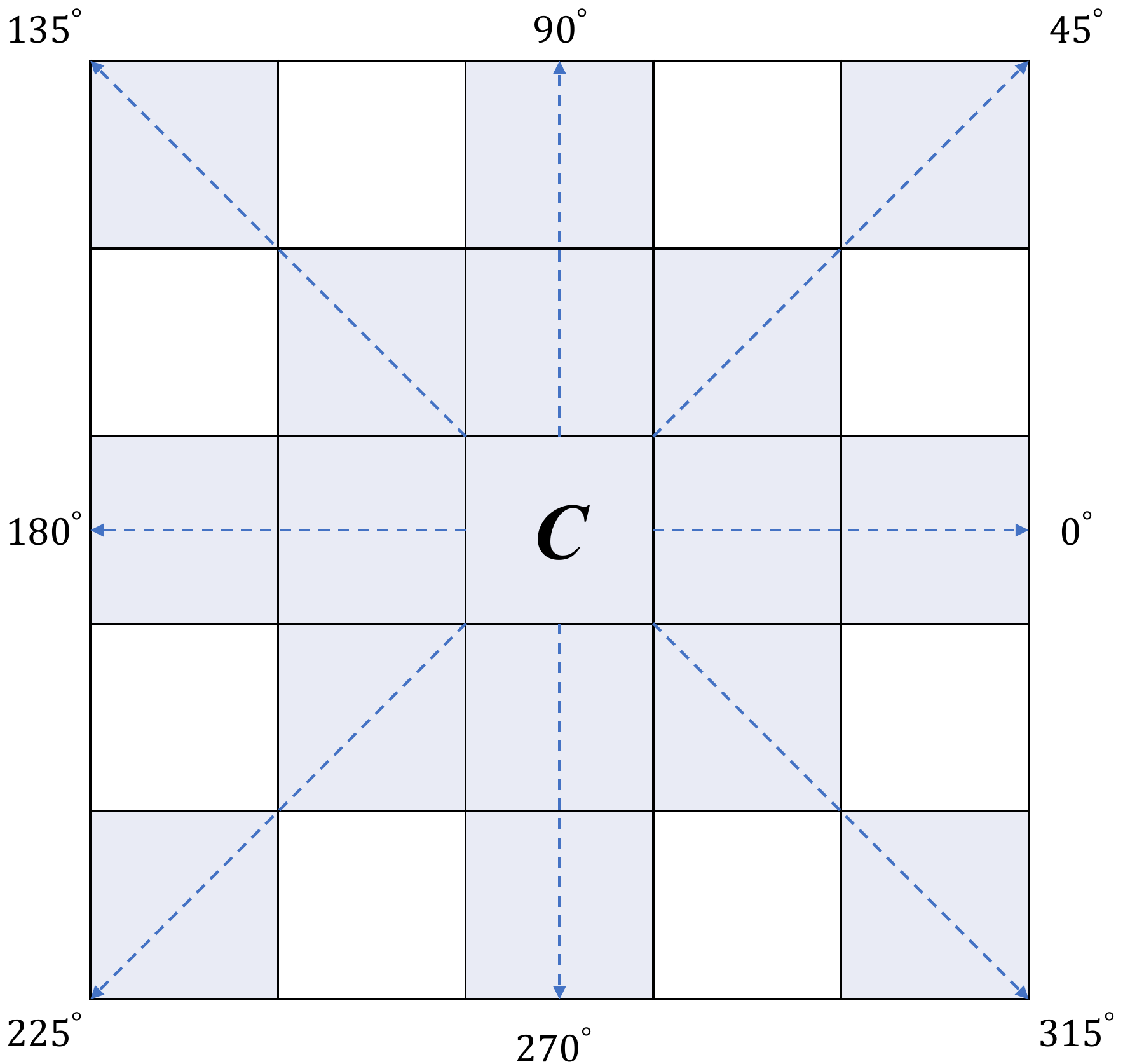}
\caption{Spatial relationship of adjacent pixels in four directions. Centered on solution cell \textbf{\textit{C}}, $0^\circ$ and $180^\circ$ are horizontal; $45^\circ$ and $225^\circ$ are anti-diagonal; $90^\circ$ and $270^\circ$ are vertical; $135^\circ$ and $315^\circ$ are diagonal.}
\label{fig_four_directions}
\end{figure}

Based on the analysis in Section \ref{section3.1.2}, we decide to calculate co-occurrence matrices \cite{haralick1973textural} in four directions (horizontal, vertical, diagonal, and anti-diagonal) for each color channel (R, G, and B) of the image. The obtained matrices are applied to analyze the spatial correlation between adjacent pixels, thus distinguishing between NIs and RIs. The similar method to find statistical data deviation between images has been well studied in the field of steganalysis \cite{fridrich2012rich,pevny2010steganalysis,sullivan2006steganalysis}. Previous steganalysis methods calculate residuals or co-occurrence matrices on the image passing through various filters. In contrast, we calculate the co-occurrence matrix of image pixels directly. This also makes it difficult to perform adversarial perturbations on co-occurrence matrices, because once perturbations are performed, the underlying statistical information of the matrices will be changed. The information loss caused by residual and filtering operation is also avoided. The co-occurrence matrix of a 2-D array $V$ can be calculated by 
\begin{equation}
\label{co-occurrence}
\begin{aligned}
&\mathbf{C}\left(\theta_{1}, \theta_{2}, \ldots, \theta_{d}\right)=\frac{1}{n}\sum_{x, y} \mathbb{I}\left(\mathbf{V}(x, y)=\theta_{1},\right. \\
&\qquad \mathbf{V}(x+\Delta x, y+\Delta y)=\theta_{2}, \ldots, \\
&\qquad \left.\mathbf{V}(x+(d-1) \Delta x, y+(d-1) \Delta y)=\theta_{d}\right),
\end{aligned}
\end{equation}
where $\mathbb{I}$(·) is the indicator function, \textit{n} is the normalization factor, $\theta_{1}, \theta_{2}, \ldots, \theta_{d}$ are the index of co-occurrence matrix ,and $\Delta x$, $\Delta y$ are the offsets for two adjacent pixels. In our practical implementation, we set the parameter \textit{d} = 2 to obtain a 2-D co-occurrence matrix, which is a 2-D histogram of pixel pair values in adjacent regions. The vertical and horizontal axes of the histogram represent the first and second values of the pair, respectively. The advantage of such operation lies in the lower complexity. The obtained co-occurrence matrices avoid the truncation operation in method \cite{fridrich2012rich}. For an image of any size with 8-bit pixel depth, our implementation will always produce a 256$\times$256 co-occurrence matrix, avoiding the need to resize the image when training and testing the network. As shown in Fig. \ref{fig_four_directions}, although there are eight possible neighbors for any pixel whose neighborhood does not exceed the image plane, only four directions need to be considered due to symmetry. Therefore, you only need to set $\Delta x$ = 1 and $\Delta y$ = 0 in (\ref{co-occurrence}) to get horizontal co-occurrence matrices, and so on in other directions.

\subsubsection{Learning features with CNN}
For any RGB image, we calculate the co-occurrence matrix in four directions for each color channel, resulting in a total of 12 matrices with size 256$\times$256. The visualization of this process is shown in Fig. \ref{fig_framework}. Most of the previous methods \cite{li2020identification,fridrich2012rich,sullivan2006steganalysis} use co-occurrence matrices to train support vector machine (SVM) classifiers. To distinguish between NIs and RIs, discriminative features need to be gleaned from the co-occurrence matrices. For this purpose, we build a feature extraction module based on CNN, which can automatically learn feature representation and is widely used in computer vision tasks such as image classification.


In this paper, our feature extraction module is built based on ResNet18 \cite{he2016deep}, where the residual block is its basic module. In the residual block, the input can forward propagate faster through the residual connections across layers. We modify the original input and output shapes in the ResNet18 network according to our task. We will get 12 matrices stacked into a tensor with size 256$\times$256$\times$12 after calculating the co-occurrence matrices, thus changing the number of the first convolution kernel from 3 to 12. The output size of the last fully connected layer of the network is also changed to 2. This output is utilized to determine whether a given image is a RI.

\section{Experiments}
\label{section4}
\subsection{Experimental setup}
\subsubsection{Image datasets}
\begin{figure}[htbp]
\centering
\includegraphics[scale=0.5]{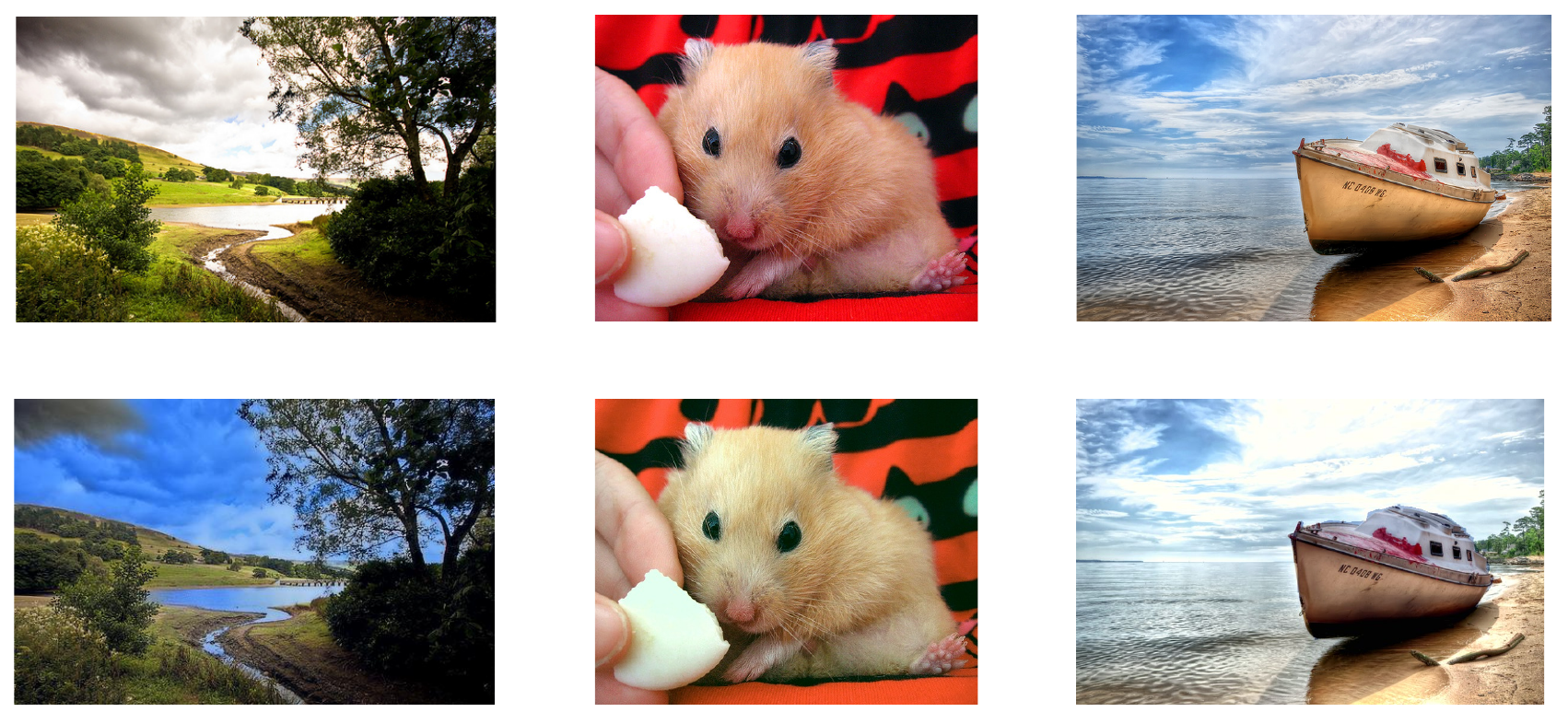}
\caption{Examples selected from the training set \textit{D}1. The first line shows NIs. The second line shows corresponding RIs generated by methods \cite{reinhard2001color,pitie2007automated,yoo2019photorealistic}, respectively.}
\label{fig_show_trainset}
\end{figure}

To train our proposed network and evaluate its effectiveness in various recoloring scenes, we utilize a variety of recoloring methods \cite{reinhard2001color,pitie2007automated,yoo2019photorealistic,lee2020deep,luan2017deep,afifi2019image,li2018closed} to create a large-scale and high-quality training set and a benchmark testing set. The training set \textit{D}1 consists of three subsets. The first subset for training consists of 19,000 NIs randomly selected from the ImageNet validation dataset \cite{russakovsky2015imagenet}, and also include their corresponding RIs, which are generated via conventional recoloring method \cite{reinhard2001color}. The ImageNet validation dataset \cite{russakovsky2015imagenet} has 1000 categories. The images in each category are randomly distributed into two parts, which are selected as style images and content images respectively. This process avoids excessive artifacts in the generated images when the style and content images are not or weakly correlated. The generation process randomly samples the style image and the content image each time and ensures that the used images are not reused. The second training subset is generated by another traditional recoloring method \cite{pitie2007automated}, with the same size as the first one. The generation process ensures that there are no overlapping images in the first and second subsets by exchanging style images and content images. The third subset for training consists of about 16,000 image pairs. In this generation process, the recoloring method \cite{yoo2019photorealistic} based on deep learning is used, and some semantic segmentation information of images in the COCO validation dataset \cite{lin2014microsoft} is also considered. Therefore, the generated images include not only the global RI but also the RI for a specific object. Fig. \ref{fig_show_trainset} shows some examples of training set.
 
\begin{figure}[htbp]
\centering
\includegraphics[scale=0.33]{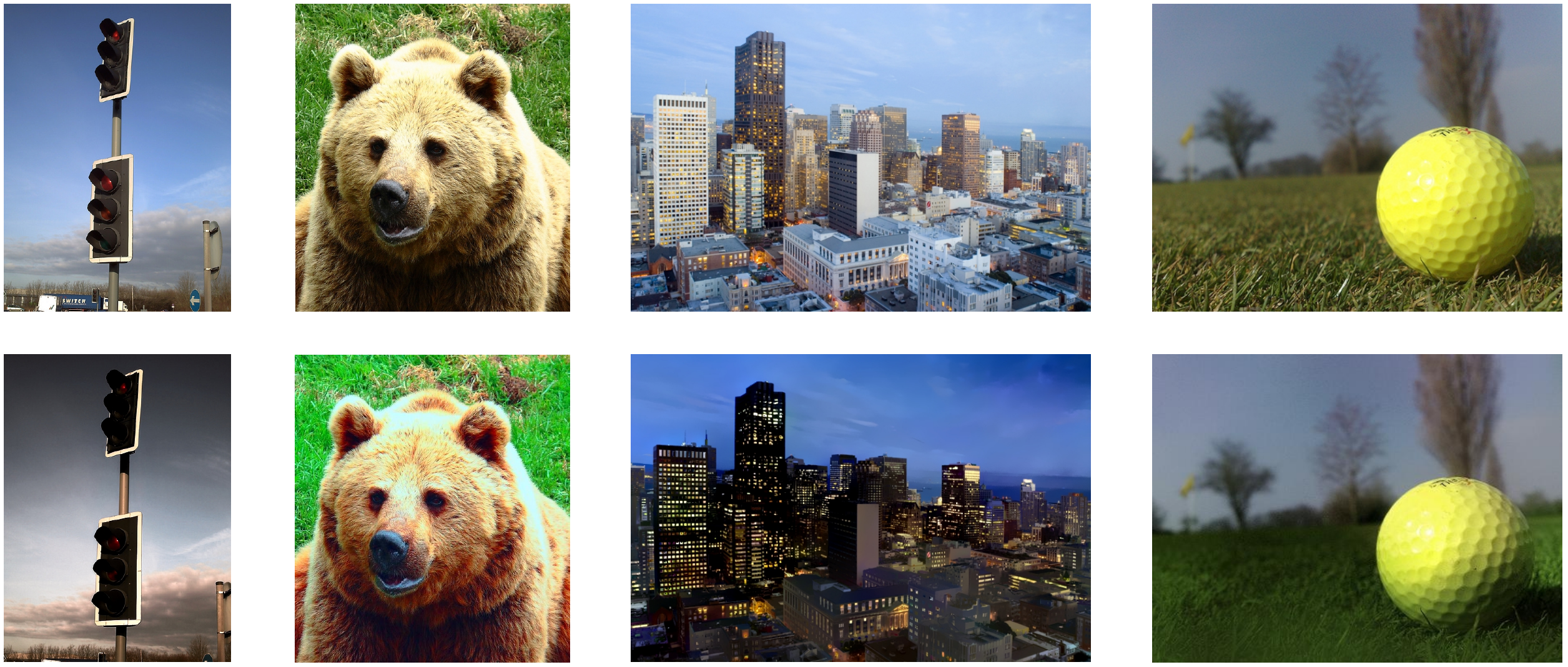}
\caption{Examples selected from the testing set \textit{D}2. The first line shows NIs. The second line shows corresponding RIs generated by methods \cite{lee2020deep,luan2017deep,afifi2019image,li2018closed}, respectively.}
\label{fig_show_testset}
\end{figure}

The testing set is divided into two challenging scenes: deep learning-based recoloring and conventional recoloring. In the deep learning-based scene, we use four deep learning-based recoloring methods \cite{lee2020deep,luan2017deep,afifi2019image,li2018closed} to generate a benchmark testing set \textit{D}2, including 240 NIs and corresponding RIs. Firstly, we randomly sample NIs from COCO validation dataset \cite{lin2014microsoft} and ImageNet validation dataset \cite{russakovsky2015imagenet} to simulate different environments. Then, each method is utilized to generates 60 recolored images. The generation process ensures that the NIs applied in each method are not repeated, and the images between the testing set and the training set are not overlapped. Fig. \ref{fig_show_testset} shows some examples of the NI and the corresponding RI in this testing set.

\begin{figure}[htbp]
\centering
\subfigure[]{
\includegraphics[width=5cm,height=4cm]{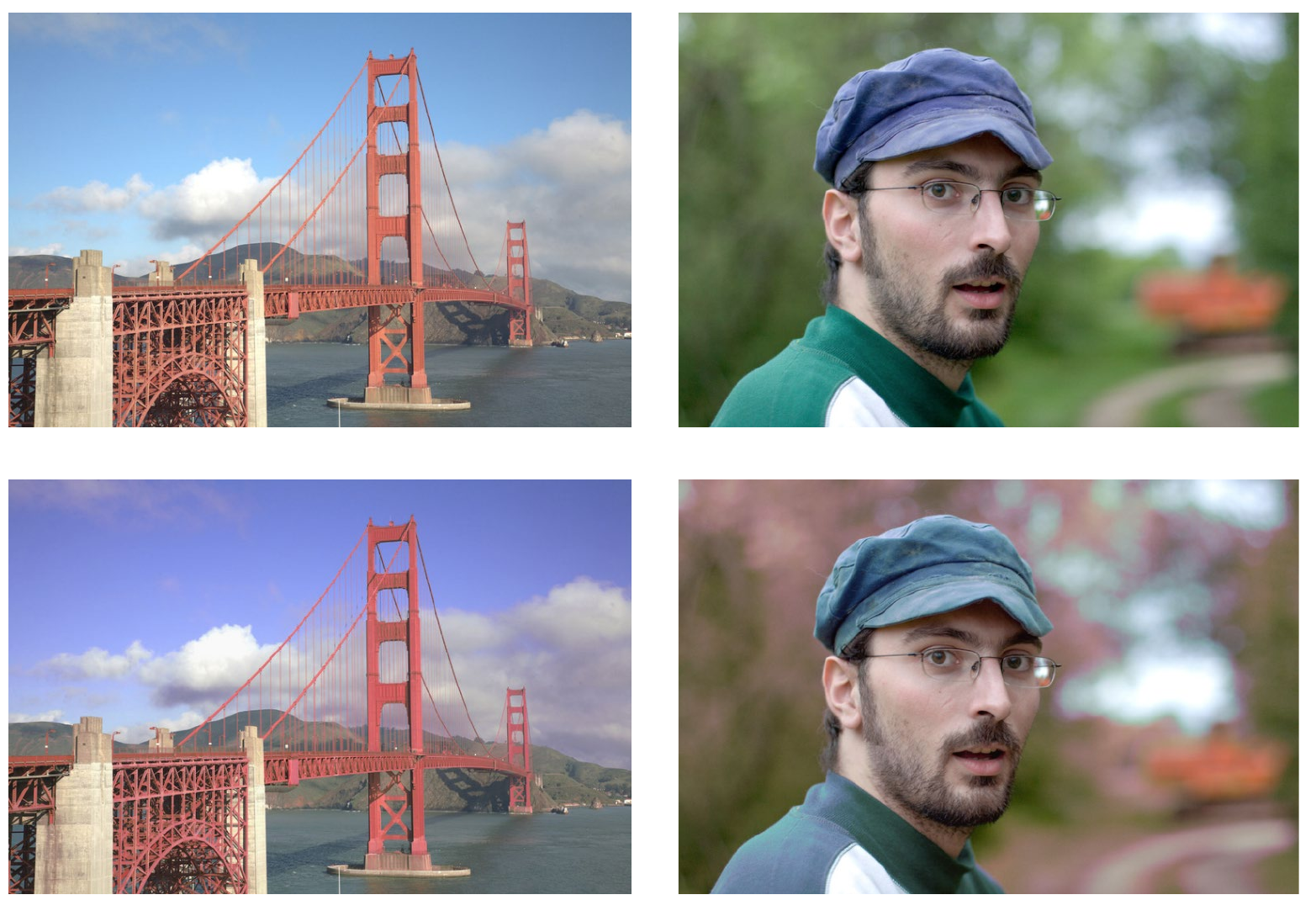}
}
\quad
\subfigure[]{
\includegraphics[width=2.5cm,height=4cm]{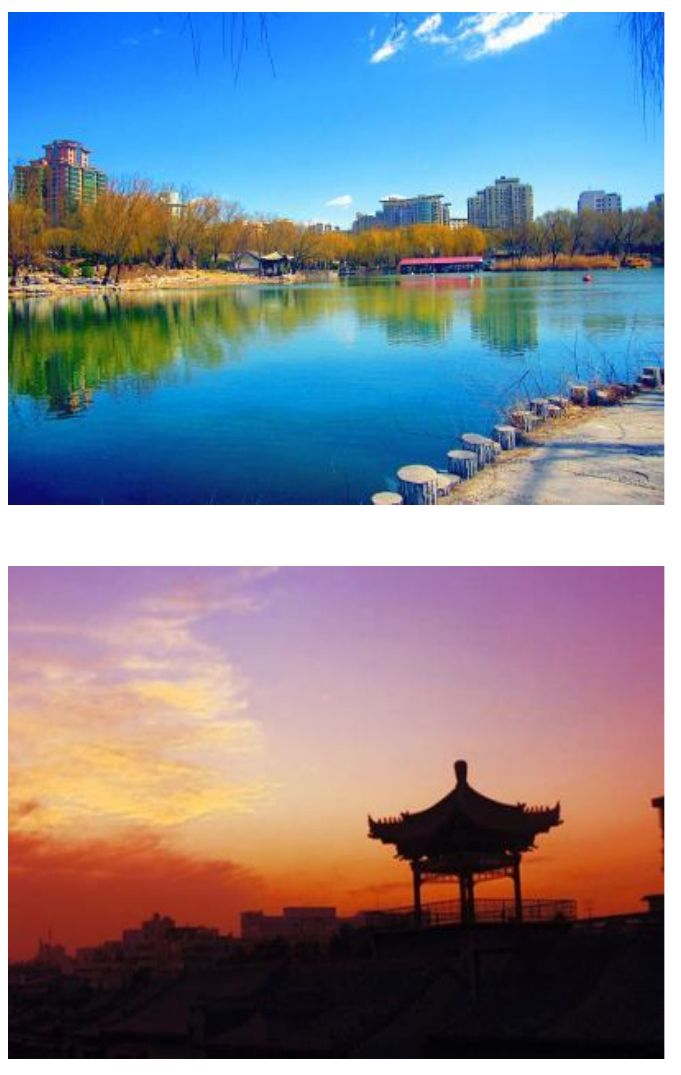}
}
\caption{Examples selected from the testing set \textit{D}3 and \textit{D}4. (a) The first line shows the NIs in \textit{D}3. The second line shows the corresponding RIs. (b) Example images randomly selected from \textit{D}4.}
\label{fig_show_conv}
\end{figure}

In conventional scenes, two recoloring testing sets released by Yan \textit{et al.} \cite{yan2018recolored} are used. Testing set \textit{D}3 includes 100 authentic images (crawled from the website) and corresponding RIs generated by a variety of conventional recoloring methods \cite{reinhard2001color,beigpour2011object,pitie2007automated,chang2015palette,pitie2007linear,grogan2015l2,an2008appprop}. The testing set \textit{D}4 contains 80 manually recolored images that were either recolored via mobile APPs or downloaded from websites. As shown in Fig. \ref{fig_show_conv}, most of the RIs in the conventional dataset look very realistic, and it is difficult for human vision to distinguish them from NIs. Refer to Table \ref{tab:summary datasets} for the basic information of all datasets.

\begin{table*}[htbp]
\centering
\scriptsize
\caption{Summary of Datasets Used in the Experiments}
\label{tab:summary datasets}
\begin{tabular}{m{1cm}<{\centering} m{2.25cm}<{\centering} m{3cm}<{\centering} m{1cm}<{\centering} m{8.5cm}<{\centering}}  
\toprule
\textbf{Dataset}             & \textbf{Category}                      & \textbf{Generation method}         & \textbf{Quantity}                   & \textbf{Note}   \\ \midrule
\multirow{3}{*}{\textit{D}1} & \multirow{2}{*}{Training set} &  \cite{reinhard2001color}  & 36,000   & Select from ImageNet validation dataset \cite{russakovsky2015imagenet}, includes NIs and corresponding RIs. \\  
    &               &  \cite{pitie2007automated} & 36,000  & Select from ImageNet validation dataset \cite{russakovsky2015imagenet}, includes NIs and corresponding RIs. \\  
    &               &  \cite{yoo2019photorealistic}  & 30,000 & Select from COCO validation dataset \cite{lin2014microsoft}, includes NIs and corresponding RIs.   \\ 
\textit{D}2 & Deep learning-based testing set &  \cite{lee2020deep,luan2017deep,afifi2019image,li2018closed}     & 480 & Select from COCO validation dataset \cite{lin2014microsoft} and ImageNet validation dataset \cite{russakovsky2015imagenet}, includes NIs and corresponding RIs.   \\ 
\textit{D}3 & Conventional testing set &  \cite{reinhard2001color,beigpour2011object,pitie2007automated,chang2015palette,pitie2007linear,grogan2015l2,an2008appprop}     & 200 & Images crawled from websites, includes NIs and corresponding RIs.   \\ 
\textit{D}4 & Conventional testing set & Manually     & 80 & Images produced by mobile APPs or downloaded from websites, only RIs are included. \\ \bottomrule
\end{tabular}
\end{table*}

\subsubsection{Implementation details}
The proposed network is implemented with the Pytorch deep learning framework \cite{paszke2019pytorch}. We adopt the Adam optimizer \cite{kingma2014adam} and set the initial learning rate as $1\times10^{-4}$. The CosineAnnealingLR \cite{loshchilov2016sgdr} method is applied to adjust the learning rate. The number of iterations of a learning rate cycle is set as 64. We also utilize L2 regularization with a weight decay of $1\times10^{-3}$. Since images of any size are calculated to obtain co-occurrence matrices with size 256$\times$256, we do not adjust the image size and the batch size is set as 128. The kernel weights are initialized with the kaiming normal distribution initialization method \cite{he2015delving}. During the training procedure, 80\% of the images in the training set \textit{D}1 are adopted for learning and updating network parameters, while the remaining 20\% are used for validation. The cross-entropy loss is used to supervise the training of network parameters. The whole network is trained 20 epochs and the data is shuffled at the beginning of each epoch. By watching the accuracy of the validation data, an early stopping strategy is adopted. If the accuracy of 5 consecutive epochs cannot be improved, the training is stopped and the model with the highest validation accuracy is saved as the final model. The training and testing are carried out on PC equipped with an NVIDIA RTX 2080 Ti GPU.

\subsubsection{Performance metrics}
For all the results reported in the paper, two commonly used metrics, i.e., accuracy and receiver operating characteristic (ROC) curve (with the area under the curve (AUC) measurement), are adopted to evaluate the performance of the proposed methods.

\subsection{Performance evaluation}
After training our network on the training set \textit{D}1, we test proposed method on a part of the images that are separated from \textit{D}1 in advance and do not overlap with training and validation instances. In these images, the accuracy reaches 95.56\% and the AUC achieves 99.22\%, indicating that our model performs well.

\begin{figure*}[htbp]
\centering
\subfigure[Natural images]{
\includegraphics[scale=0.35]{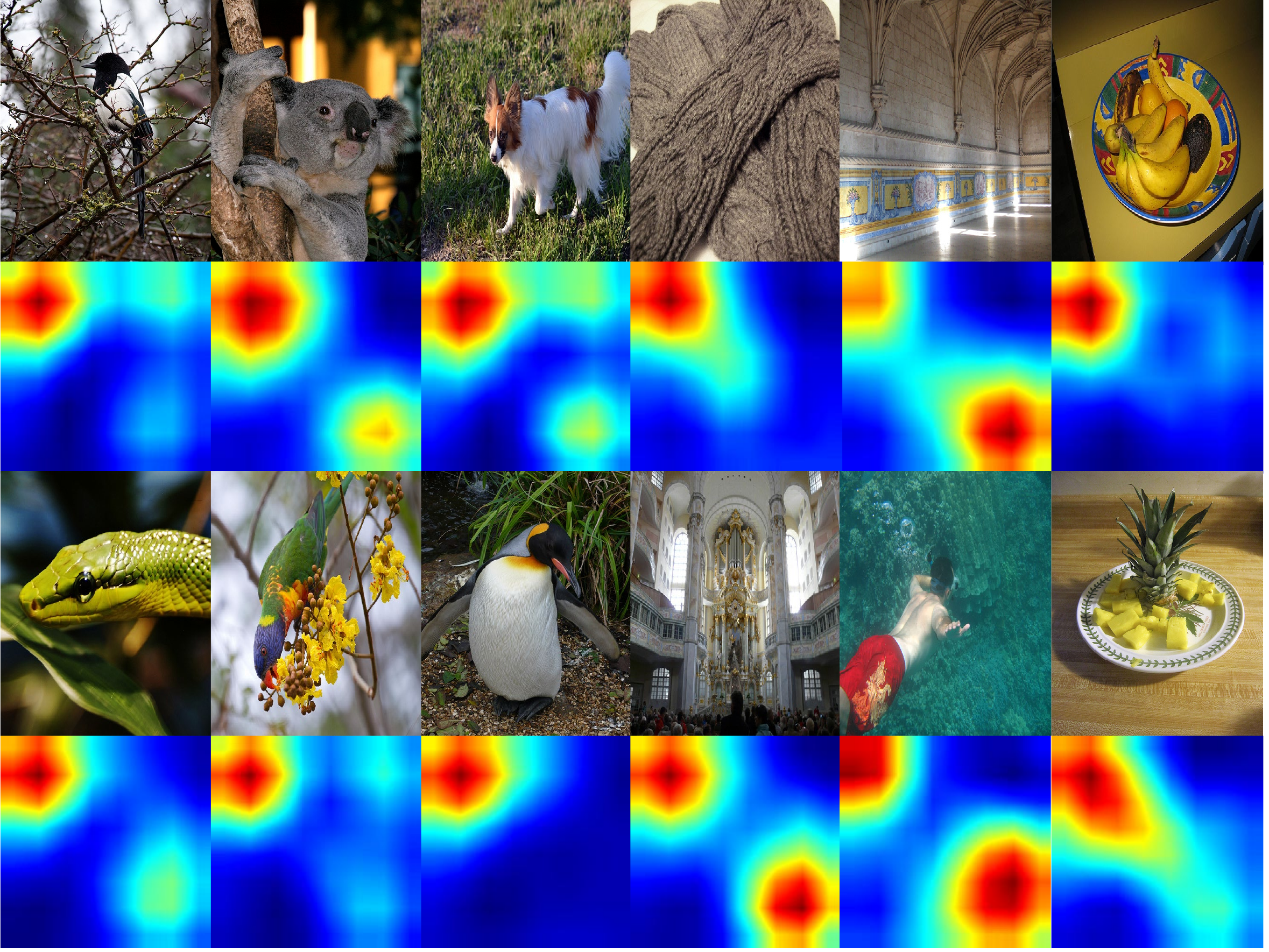}
}
\quad
\subfigure[Recolored images]{
\includegraphics[scale=0.35]{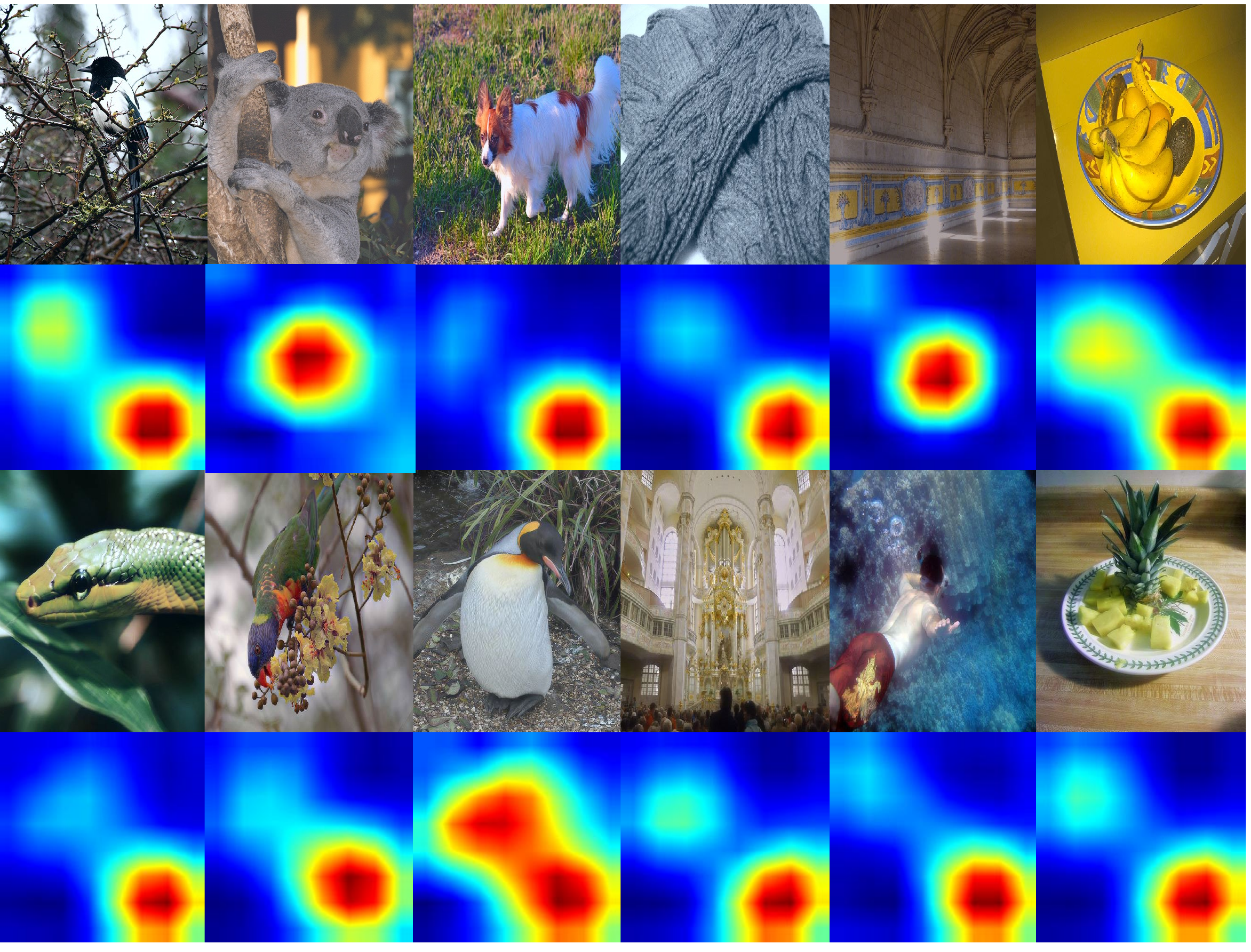}
}
\caption{The difference in heatmap between NIs (a) and their corresponding RIs (b).}
\label{fig_heatmap}
\end{figure*}

Fig. \ref{fig_heatmap} shows the heatmaps of the last convolution layer in our network using Grad-CAM$++$ \cite{chattopadhay2018grad} which can provide better visual interpretability for our CNN model predictions. We can see that although NIs and corresponding RIs look familiar, their heatmaps are quite different. The upper left corner regions of the heatmaps of the NIs are predominantly red (that is, negative correlation).  On contrast, the upper left corner regions of the heatmaps of the corresponding RI are predominantly blue (that is, positive correlation). The visualization results of the heatmap show that our network can further learn from the extracted spatial correlation feature set, and can distinguish between NIs and RIs.

To further evaluate the practicability of our network and its performance in dealing with various recolor challenges, we will conduct experiments in deep learning-based recoloring and conventional recoloring scenes, respectively.

\subsubsection{Deep learning-based recoloring scene}
In this case, we evaluate the detection performance of the proposed network in the recoloring scene based on deep learning, i.e., the RIs are all generated using deep learning-based methods. We utilize the network trained above to examine its performance on the testing set \textit{D}2. Although the images in \textit{D}2 and \textit{D}1 do not overlap at all and are generated using completely different recoloring methods, the proposed network can still achieves the accuracy of 85.21\% and the AUC of 94.09\%.  Table \ref{tab:results with different testing set} shows the detection accuracy and AUC of the proposed method with various subsets. It reveals that our model may have learned the common spatial correlation features of RIs, and have good generalization.

\begin{table}[htbp]
\centering
\scriptsize
\caption{Results of Proposed Method on Different Testing Datasets}
\label{tab:results with different testing set}
\begin{tabular}{ m{3.0cm}<{\centering} m{1.5cm}<{\centering} m{1.25cm}<{\centering} }
    \toprule
        \textbf{Testing set} & \textbf{Accuracy (\%)} & \textbf{AUC (\%)} \\
    \midrule
         \textit{D}1 & 95.56  & 99.22 \\
         \textit{D}2 & 85.21  & 94.09 \\
         \textit{D}3 & 82.50  & 91.21 \\
         \textit{D}4 & 70.00  & --- \\
    \bottomrule
\end{tabular}
\end{table}

\subsubsection{Conventional recoloring scene}
In this case, we apply the previously trained network to examine the effectiveness of the proposed method for the conventional recoloring scene on the testing sets \textit{D}3 and \textit{D}4. The conventional recoloring methods have the characteristics of low computational cost and good interpretability. The manual recoloring methods can achieve a more realistic recoloring effect, and the details are also handled better. On \textit{D}3, the accuracy of our trained model reaches 82.50\% and the AUC achieves 91.21\%. On \textit{D}4, the detection accuracy is 70.00\%. 
It is 14.50\% higher than the human vision accuracy of 55.50\%\footnote{The data is cited directly from the Section IV-A of Yan \textit{et al.} \cite{yan2018recolored}.}. The accuracy on \textit{D}4 is a bit lower than that on \textit{D}3. This is possibly because of that images in \textit{D}4 are affected by human prior knowledge, which can produce the most realistic effect with the lowest cost of perturbing the spatial correlation of pixels in the images and our training data do not cover these images. In conclusion, the detection results on \textit{D}3 and \textit{D}4 show that our model has a good capability to deal with conventional recoloring scenes composed of conventional methods and manual recoloring with automated tools.

\subsection{Ablation study}
In this subsection, we conduct an ablation study to show the superiority of  the proposed method over its variants. This experiment is conducted on the testing set \textit{D}2. We first modify the preprocessing module, the CNN architecture, and the used training set, respectively. Then, we train different models with different settings to detect the images on \textit{D}2. Finally, according to the results, the influence of different settings on the model performance is analyzed.

\subsubsection{On preprocessing module}
\begin{figure}[htbp]
\centering
\includegraphics[scale=0.5]{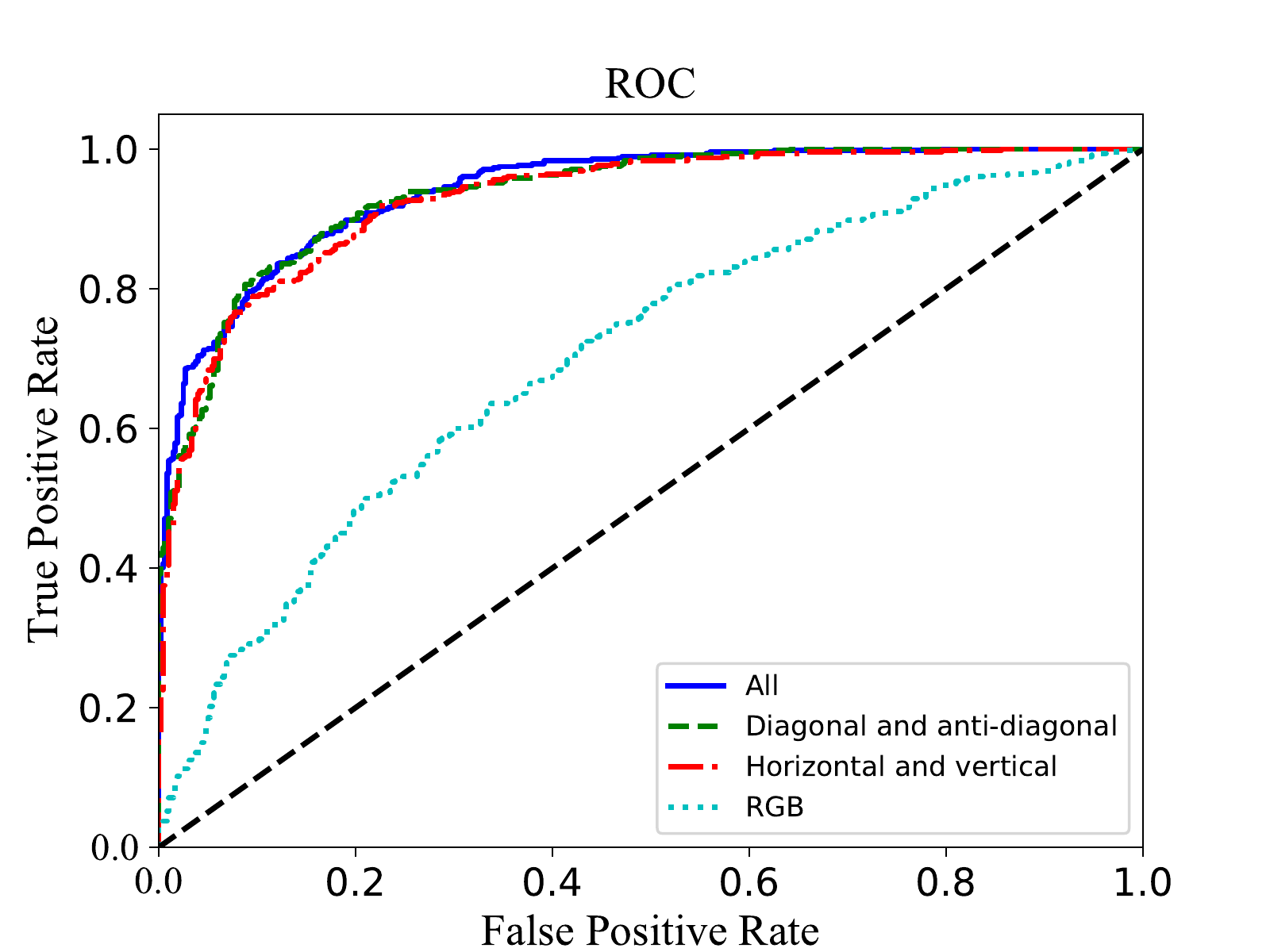}
\caption{The ROC curve of the proposed methods with respect to different inputs on testing set \textit{D}2.}
\label{fig_ablation_co_oc}
\end{figure}

\begin{figure}[htbp]
\centering
\subfigure[]{
\includegraphics[scale=0.25]{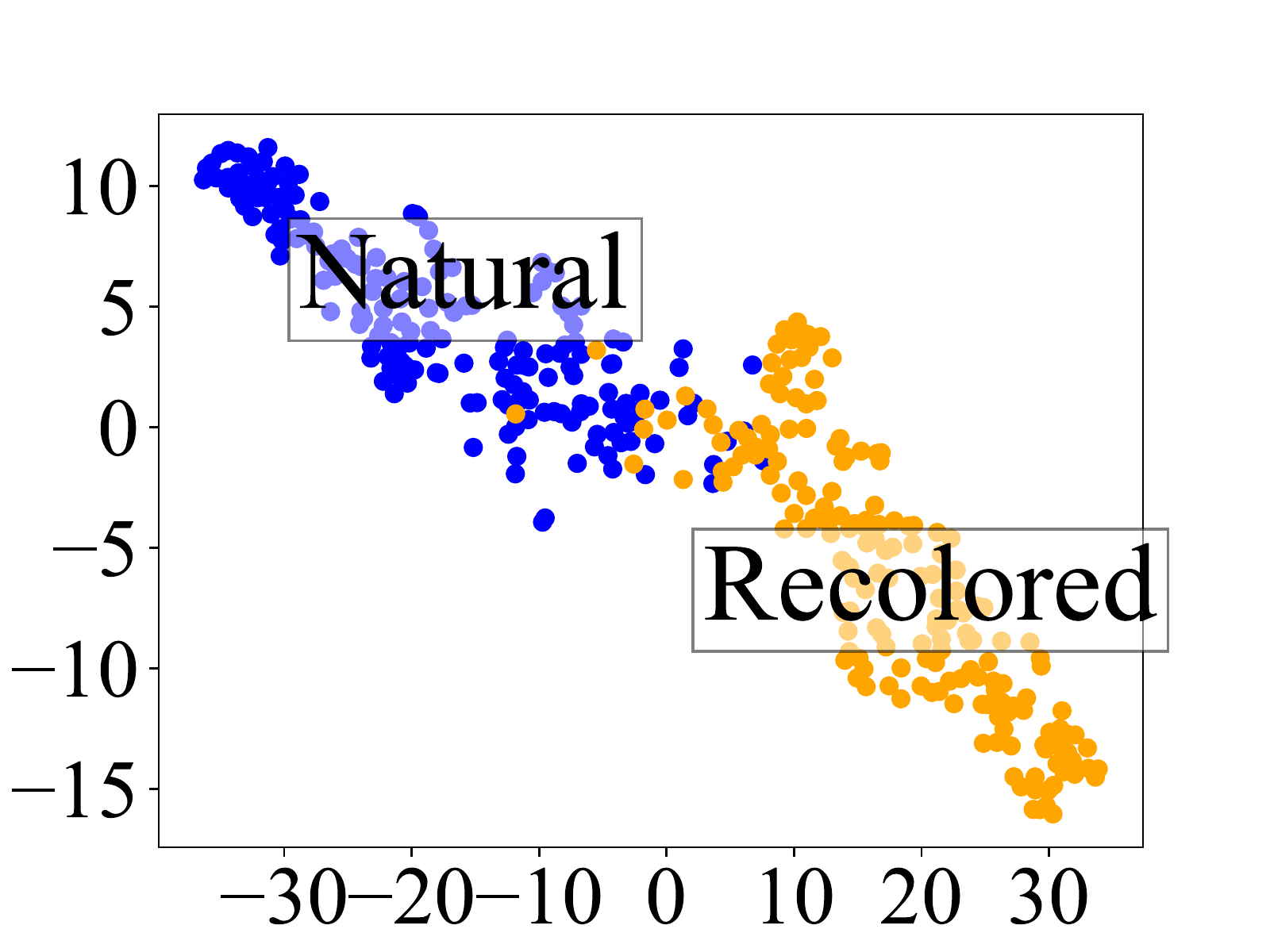}
}
\subfigure[]{
\includegraphics[scale=0.25]{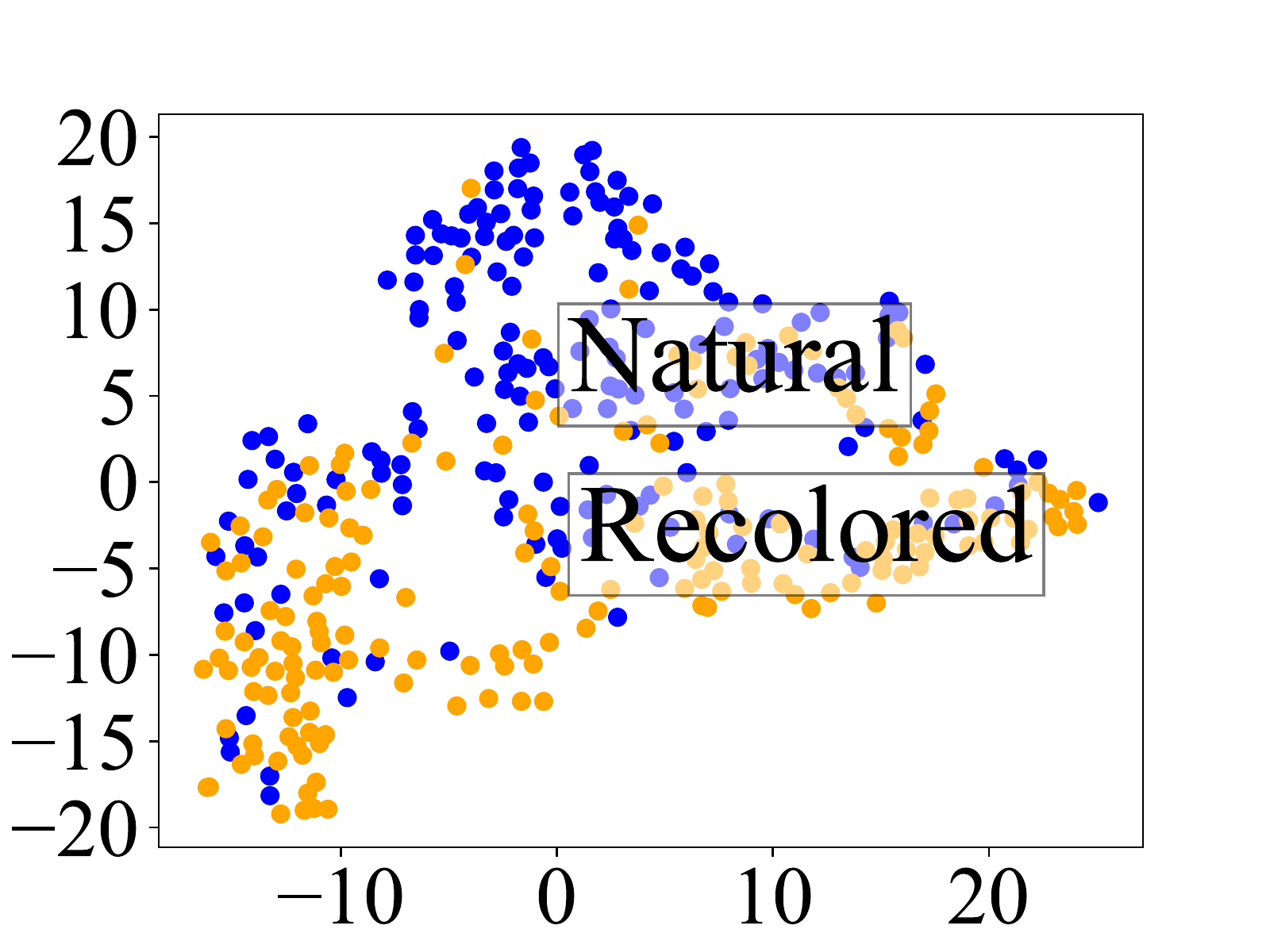}
}
\caption{Visualization difference between the proposed method and baseline. Blue points representing the NIs and the orange points representing the RIs. (a) The input of the network are the feature sets. (b) The input of the network are original RGB images.}
\label{fig_visulation}
\end{figure}

As shown in Fig. \ref{fig_framework}, the preprocessing module extracts the spatial correlation between adjacent pixels in four directions by calculating the co-occurrence matrix. Here, we set the baseline to not extract spatial correlation. We directly feed the original RGB images into CNN architecture, and adjust the size of all images to 224$\times$224 for the convenience of training. In addition, we also take the feature set that only calculates the spatial correlation of two directions as the input of the network, i.e., the horizontal and vertical directions are divided into a group, and the diagonal and anti-diagonal are in another group. The comparison results of the above three variants and the complete inputs are summarized in Table \ref{tab:results with different inputs}. When calculating the spatial correlation in the four directions, our model achieves the highest accuracy and AUC for the same training and testing dataset. Fig. \ref{fig_ablation_co_oc} shows ROC curves for different inputs. Note that when only the spatial correlation of two directions is considered, the discriminative effect is significantly improved compared to the baseline. Fig. \ref{fig_visulation} shows the feature space of the penultimate layer in our network using t-SNE visualization \cite{van2008visualizing}. In Fig. \ref{fig_visulation} (a), the inputs of our network are the feature sets that indicate the spatial correlation. Blue points represent NIs, and orange points represent RIs. It can be seen that the blue and the orange points are clustered into two clusters with a clear boundary. On the contrary, the blue and orange points are blended together in Fig. \ref{fig_visulation} (b). It shows that when the inputs are original RGB images, the network cannot distinguish between NIs and RIs well. Therefore, spatial correlation is indeed the discriminative feature. It is necessary to consider the spatial correlation in four directions for better performance. 

\begin{table}[htbp]
\centering
\scriptsize
\caption{Result of Proposed Method with Different Inputs}
\label{tab:results with different inputs}
\begin{tabular}{ m{3.0cm}<{\centering} m{1.5cm}<{\centering} m{1.25cm}<{\centering} }
    \toprule
        \textbf{Inputs} & \textbf{Accuracy (\%)} & \textbf{AUC (\%)} \\ \midrule
        All & \textbf{85.21}  & \textbf{94.09} \\
        Diagonal and anti-diagonal & 85.00  & 93.56 \\
         Horizontal and vertical & 83.96 & 92.87 \\
        RGB & 65.21 & 70.10 \\ \bottomrule
\end{tabular}
\end{table}

\subsubsection{On CNN architecture}
\begin{figure}[htbp]
\centering
\includegraphics[scale=0.5]{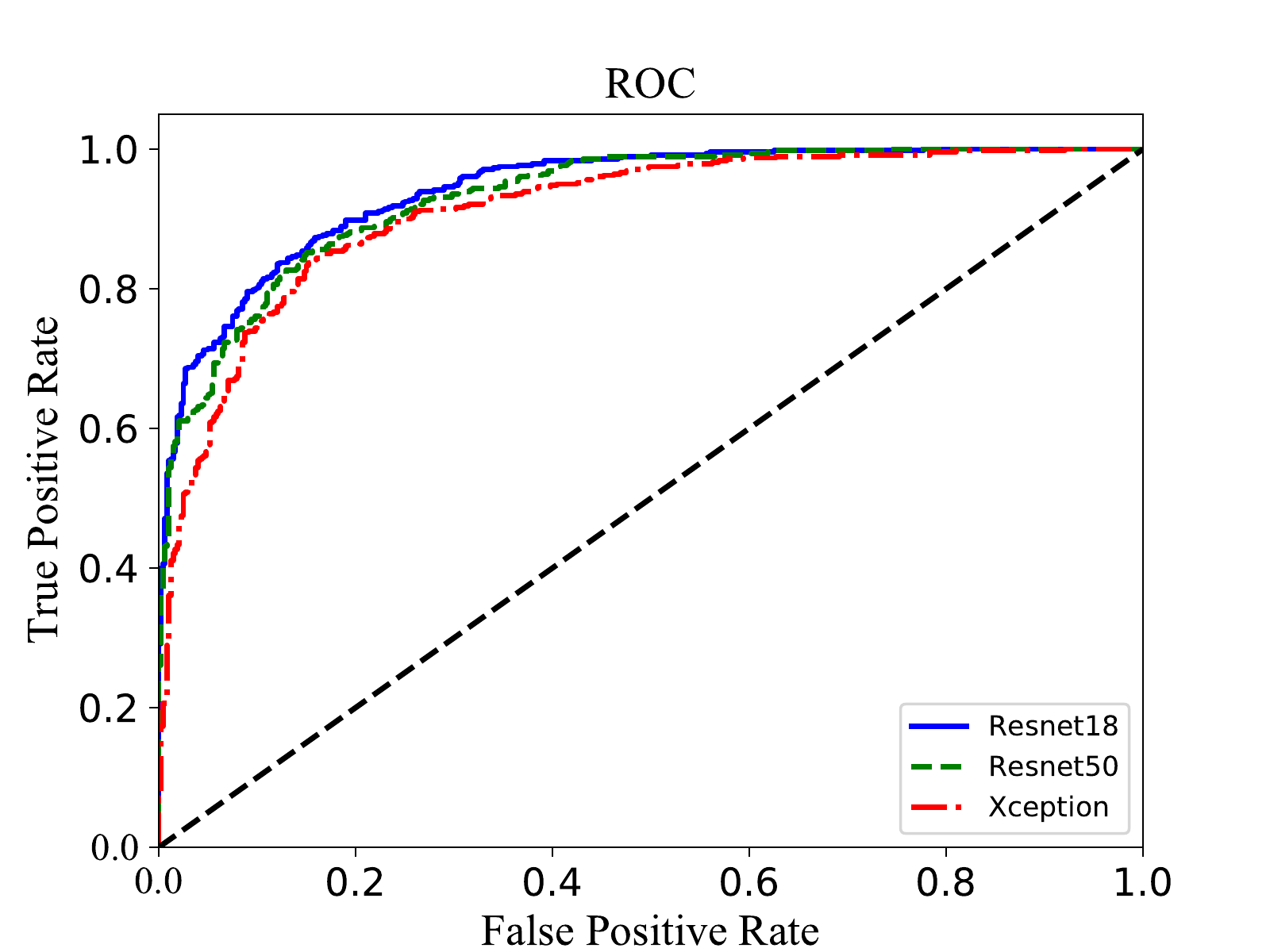}
\caption{The ROC Curve of the proposed methods with respect to different CNN architectures on testing set \textit{D}2.}
\label{fig_ablation_backbone}
\end{figure}

In this section, we evaluate the impact of different CNN architectures on the results. In previous experiments, we feed the extracted spatial correlation features to Resnet18 \cite{he2016deep} for further learning. Here, we evaluate our method on different well-known CNN architectures: Resnet18 \cite{he2016deep}, Resnet50 \cite{he2016deep} and Xception \cite{chollet2017xception}. The results are shown in Table \ref{tab:results with different cnn} and ROC curves are shown in Fig. \ref{fig_ablation_backbone}. Although Resnet50 and Xception have a deeper network structure and greater learning capabilities, the accuracy of the three CNN architectures is not much different, and Resnet18 is still the best. This may be because complex networks are overfitted and learn some specific features of a particular recoloring method, leading to unsatisfactory applicability. Therefore, our network using Resnet18 avoids overfitting and saves training time.

\begin{table}[htbp]
\centering
\scriptsize
\caption{Results of Proposed Method Based on Different Networks}
\label{tab:results with different cnn}
\begin{tabular}{ m{3.0cm}<{\centering} m{1.5cm}<{\centering} m{1.25cm}<{\centering} }
    \toprule
        \textbf{CNN networks} & \textbf{Accuracy (\%)} & \textbf{AUC (\%)} \\
    \midrule
        Resnet18 \cite{he2016deep} & \textbf{85.21}  & \textbf{94.09} \\
         Resnet50 \cite{he2016deep} & \textbf{85.21} & 93.11 \\
        Xception \cite{chollet2017xception} & 84.17  & 91.15 \\
    \bottomrule
\end{tabular}
\end{table}

\subsubsection{On training set used}
\begin{figure}[htbp]
\centering
\includegraphics[scale=0.5]{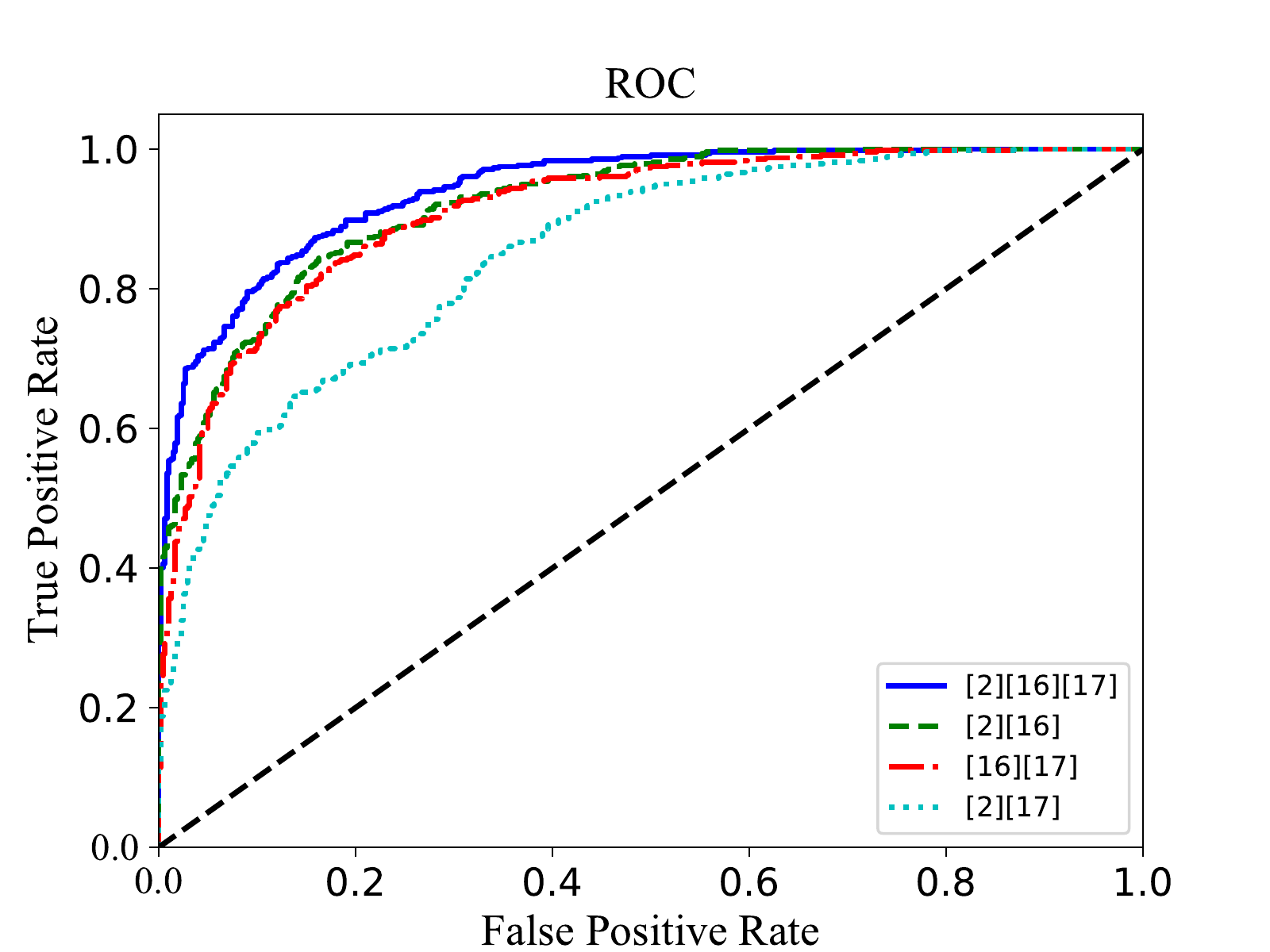}
\caption{The ROC Curve of the proposed methods with respect to different training sets on testing set \textit{D}2.}
\label{fig_ablation_train_set}
\end{figure}

The networks we applied above are trained on training set \textit{D}1. In this section, we try to utilize only the subsets of \textit{D}1 to train the network, and verify the impact of the training set on the proposed method. Since \textit{D}1 has three subsets, each of which is generated using a recoloring method, we apply two of the three subsets to train the proposed network and then evaluate it on testing set \textit{D}2. The reason why we do not consider applying only one subset is to prevent the network from learning specific features of a particular recoloring method, rather than dealing with scenes consisting of various recoloring methods. Table \ref{tab:results with different training set} and Fig. \ref{fig_ablation_train_set} show the detection results and ROC curves of the proposed method under different subsets, respectively. Although the results of using subsets are lower than the case with whole \textit{D}1, our method still performs well with the accuracy of 75.21\% and the AUC of 79.52\% at least, which indicate that our approach indeed learns the common spatial correlation features of RIs.

\begin{table}[htbp]
\centering
\scriptsize
\caption{Results of Proposed Method with Different Training Datasets}
\label{tab:results with different training set}
\begin{tabular}{ m{3.0cm}<{\centering} m{1.5cm}<{\centering} m{1.25cm}<{\centering} }
    \toprule
        \textbf{Training set} & \textbf{Accuracy (\%)} & \textbf{AUC (\%)} \\
    \midrule
         \cite{reinhard2001color,pitie2007automated,yoo2019photorealistic} & \textbf{85.21}  & \textbf{94.09} \\
          \cite{reinhard2001color,pitie2007automated} & 83.96 & 92.04 \\
         \cite{pitie2007automated,yoo2019photorealistic} & 82.71  & 91.06 \\
        \cite{reinhard2001color,yoo2019photorealistic} & 75.21  & 79.52 \\
    \bottomrule
\end{tabular}
\end{table}

\subsection{Comparisons with the state-of-the-art methods}
In this experiment, we compare our method with two state of-the-art methods \cite{yan2018recolored,goljan2014rich}. Yan \textit{et al.}'s method \cite{yan2018recolored} used the original RGB image with its DI and IM to detect RIs. Rich-features-based method \cite{goljan2014rich} first extracted SCRMQ1 features and then trained an SVM for classification. Yan \textit{et al.} \cite{yan2018recolored} only released the testing sets \textit{D}3 and \textit{D}4, but without the training sets they generated using methods \cite{reinhard2001color,pitie2007automated,beigpour2011object}. In order to compare as fairly as possible, we train our network using subsets of \textit{D}1 generated by methods \cite{reinhard2001color,pitie2007automated}. Table \ref{tab:Comparisons} shows the comparison results\footnote{ Results of Yan \textit{et al.} \cite{yan2018recolored} and Rich-features-based method \cite{goljan2014rich} are cited directly from  \cite{yan2018recolored}.}. The accuracy of our work on \textit{D}3 is slightly lower than that of Yan \textit{et al.} \cite{yan2018recolored}. Our approach achieves comparable results while using less training data. On the contrary, for the manual dataset \textit{D}4, our method achieves the highest accuracy for completely unaware recoloring scenes, which shows that our method is more general.

\begin{table}[htbp]
\centering
\scriptsize
\caption{Comparisons with State-of-the-Art Methods}
\label{tab:Comparisons}
\begin{tabular}{ m{1cm}<{\centering} m{3.5cm}<{\centering} m{1.5cm}<{\centering} }  \toprule
 \textbf{Test data}                    & \textbf{Methods}                                         & \textbf{Accuracy (\%)} \\ \midrule
\multirow{3}{*}{\textit{D}3} & Yan \textit{et al.} \cite{yan2018recolored}      & \textbf{83.50}        \\  
                             & Rich-features-based method \cite{goljan2014rich} & 69.00        \\ 
                             & Ours (using subsets trained)                     & 77.00        \\ 
                            
\multirow{3}{*}{\textit{D}4} & Yan \textit{et al.} \cite{yan2018recolored}      & 65.00        \\  
                             & Rich-features-based method \cite{goljan2014rich} & 37.50        \\  
                             & Ours (using subsets trained)                     & \textbf{73.75}        \\ 
                             \bottomrule
\end{tabular}
\end{table}

\section{Conclusion}
In this paper, we propose a deep learning approach for RI detection by using spatial correlation, which exhibits the generic detection capability for both conventional and deep learning-based recoloring. Through theoretical and numerical analysis, we observe that although NIs and RIs are visually difficult to distinguish, their spatial correlation between adjacent pixels on each color component is quite discriminative. Based on such revealed prior knowledge, we extract the correlation by calculating the co-occurrence matrices. The CNN is then applied to such matrices for learning informative representations, thus detecting RIs. We verify the rationality of the whole network through several ablation experiments. The state-of-the-art performance and generalization on the datasets generated by various recoloring methods indicate the effectiveness of our method. We hope that our findings on the spatial correlation of adjacent pixels will provide useful insight for recoloring detection tasks. In the future, we will focus on improving detection performance by exploring better feature encoding methods and designing a more effective network architecture.
\label{section5}
\bibliographystyle{IEEEtran}
\bibliography{IEEEabrv,references}

\begin{thebibliography}{10}
\providecommand{\url}[1]{#1}
\csname url@samestyle\endcsname
\providecommand{\newblock}{\relax}
\providecommand{\bibinfo}[2]{#2}
\providecommand{\BIBentrySTDinterwordspacing}{\spaceskip=0pt\relax}
\providecommand{\BIBentryALTinterwordstretchfactor}{4}
\providecommand{\BIBentryALTinterwordspacing}{\spaceskip=\fontdimen2\font plus
\BIBentryALTinterwordstretchfactor\fontdimen3\font minus
  \fontdimen4\font\relax}
\providecommand{\BIBforeignlanguage}[2]{{%
\expandafter\ifx\csname l@#1\endcsname\relax
\typeout{** WARNING: IEEEtran.bst: No hyphenation pattern has been}%
\typeout{** loaded for the language `#1'. Using the pattern for}%
\typeout{** the default language instead.}%
\else
\language=\csname l@#1\endcsname
\fi
#2}}
\providecommand{\BIBdecl}{\relax}
\BIBdecl

\bibitem{nightingale2017can}
S.~J. Nightingale, K.~A. Wade, and D.~G. Watson, ``Can people identify original
  and manipulated photos of real-world scenes?'' \emph{Cogn.Res.: Princ.
  Implications}, vol.~2, no.~1, pp. 1--21, 2017.

\bibitem{reinhard2001color}
E.~Reinhard, M.~Adhikhmin, B.~Gooch, and P.~Shirley, ``Color transfer between
  images,'' \emph{IEEE Comput. Graph. Appl.}, vol.~21, no.~5, pp. 34--41, 2001.

\bibitem{verdoliva2020media}
L.~Verdoliva, ``Media forensics and deepfakes: an overview,'' \emph{IEEE J.
  Sel. Topics Signal Process.}, vol.~14, no.~5, pp. 910--932, 2020.

\bibitem{farid2009image}
H.~Farid, ``Image forgery detection,'' \emph{IEEE Signal Process. Mag.},
  vol.~26, no.~2, pp. 16--25, 2009.

\bibitem{stamm2013information}
M.~C. Stamm, M.~Wu, and K.~R. Liu, ``Information forensics: An overview of the
  first decade,'' \emph{IEEE Access}, vol.~1, pp. 167--200, 2013.

\bibitem{wang2017hybrid}
J.~Wang, S.~Lian, and Y.-Q. Shi, ``Hybrid multiplicative multi-watermarking in
  dwt domain,'' \emph{Multidimensional Syst. Signal Process.}, vol.~28, no.~2,
  pp. 617--636, 2017.

\bibitem{yerushalmy2011digital}
I.~Yerushalmy and H.~Hel-Or, ``Digital image forgery detection based on lens
  and sensor aberration,'' \emph{Int. J. Comput. Vis.}, vol.~92, no.~1, pp.
  71--91, 2011.

\bibitem{popescu2005exposing}
A.~C. Popescu and H.~Farid, ``Exposing digital forgeries by detecting traces of
  resampling,'' \emph{IEEE Trans. Signal Process.}, vol.~53, no.~2, pp.
  758--767, 2005.

\bibitem{cozzolino2019noiseprint}
D.~Cozzolino and L.~Verdoliva, ``Noiseprint: A cnn-based camera model
  fingerprint,'' \emph{IEEE Trans. Inf. Forensics security}, vol.~15, pp.
  144--159, 2019.

\bibitem{lukavs2003estimation}
J.~Luk{\'a}{\v{s}} and J.~Fridrich, ``Estimation of primary quantization matrix
  in double compressed jpeg images,'' in \emph{Proc. Digital Forensic Research
  Workshop}, 2003, pp. 5--8.

\bibitem{cao2014contrast}
G.~Cao, Y.~Zhao, R.~Ni, and X.~Li, ``Contrast enhancement-based forensics in
  digital images,'' \emph{IEEE Trans. Inf. Forensics Security}, vol.~9, no.~3,
  pp. 515--525, 2014.

\bibitem{li2018fast}
Y.~Li and J.~Zhou, ``Fast and effective image copy-move forgery detection via
  hierarchical feature point matching,'' \emph{IEEE Trans. Inf. Forensics
  Security}, vol.~14, no.~5, pp. 1307--1322, 2018.

\bibitem{matern2019gradient}
F.~Matern, C.~Riess, and M.~Stamminger, ``Gradient-based illumination
  description for image forgery detection,'' \emph{IEEE Trans. Inf. Forensics
  Security}, vol.~15, pp. 1303--1317, 2019.

\bibitem{li2019localization}
H.~Li and J.~Huang, ``Localization of deep inpainting using high-pass fully
  convolutional network,'' in \emph{Proc. IEEE/CVF Int. Conf. Comput. Vis.
  (ICCV)}, 2019, pp. 8301--8310.

\bibitem{yan2018recolored}
Y.~Yan, W.~Ren, and X.~Cao, ``Recolored image detection via a deep
  discriminative model,'' \emph{IEEE Trans. Inf. Forensics Security}, vol.~14,
  no.~1, pp. 5--17, 2018.

\bibitem{pitie2007automated}
F.~Piti{\'e}, A.~C. Kokaram, and R.~Dahyot, ``Automated colour grading using
  colour distribution transfer,'' \emph{Comput. Vis. Image Understand.}, vol.
  107, no. 1-2, pp. 123--137, 2007.

\bibitem{yoo2019photorealistic}
J.~Yoo, Y.~Uh, S.~Chun, B.~Kang, and J.-W. Ha, ``Photorealistic style transfer
  via wavelet transforms,'' in \emph{Proc. IEEE/CVF Int. Conf. Comput. Vis.
  (ICCV)}, 2019, pp. 9036--9045.

\bibitem{lee2020deep}
J.~Lee, H.~Son, G.~Lee, J.~Lee, S.~Cho, and S.~Lee, ``Deep color transfer using
  histogram analogy,'' \emph{The Visual Comput.}, vol.~36, no.~10, pp.
  2129--2143, 2020.

\bibitem{luan2017deep}
F.~Luan, S.~Paris, E.~Shechtman, and K.~Bala, ``Deep photo style transfer,'' in
  \emph{Proc. IEEE Conf. Comput. Vis. Pattern Recognit. (CVPR)}, 2017, pp.
  4990--4998.

\bibitem{afifi2019image}
M.~Afifi, B.~L. Price, S.~Cohen, and M.~S. Brown, ``Image recoloring based on
  object color distributions,'' in \emph{Eurographics (Short Papers)}, 2019,
  pp. 33--36.

\bibitem{li2018closed}
Y.~Li, M.-Y. Liu, X.~Li, M.-H. Yang, and J.~Kautz, ``A closed-form solution to
  photorealistic image stylization,'' in \emph{Proc. Eur. Conf. Comput. Vis.
  (ECCV)}, 2018, pp. 453--468.

\bibitem{pitie2005n}
F.~Pitie, A.~C. Kokaram, and R.~Dahyot, ``N-dimensional probability density
  function transfer and its application to color transfer,'' in \emph{Tenth
  IEEE Int. Conf. Comput. Vis. (ICCV'05)}, vol.~2, 2005, pp. 1434--1439.

\bibitem{he2019progressive}
M.~He, J.~Liao, D.~Chen, L.~Yuan, and P.~V. Sander, ``Progressive color
  transfer with dense semantic correspondences,'' \emph{ACM Trans. Graph.
  (TOG)}, vol.~38, no.~2, pp. 1--18, 2019.

\bibitem{tai2005local}
Y.-W. Tai, J.~Jia, and C.-K. Tang, ``Local color transfer via probabilistic
  segmentation by expectation-maximization,'' in \emph{IEEE Comput. Society
  Conf. Comput. Vis. Pattern Vis. (CVPR'05)}, vol.~1, 2005, pp. 747--754.

\bibitem{levin2004colorization}
A.~Levin, D.~Lischinski, and Y.~Weiss, ``Colorization using optimization,''
  \emph{ACM Trans. Graph. (TOG)}, vol.~23, no.~3, pp. 689--694, 2004.

\bibitem{qu2006manga}
Y.~Qu, T.-T. Wong, and P.-A. Heng, ``Manga colorization,'' \emph{ACM Trans.
  Graph. (TOG)}, vol.~25, no.~3, pp. 1214--1220, 2006.

\bibitem{an2008appprop}
X.~An and F.~Pellacini, ``Appprop: all-pairs appearance-space edit
  propagation,'' \emph{ACM Trans. Graph. (TOG)}, vol.~27, no.~3, pp. 1--9,
  2008.

\bibitem{xu2009efficient}
K.~Xu, Y.~Li, T.~Ju, S.-M. Hu, and T.-Q. Liu, ``Efficient affinity-based edit
  propagation using kd tree,'' \emph{ACM Trans. Graph. (TOG)}, vol.~28, no.~5,
  pp. 1--6, 2009.

\bibitem{chen2014sparse}
X.~Chen, D.~Zou, J.~Li, X.~Cao, Q.~Zhao, and H.~Zhang, ``Sparse dictionary
  learning for edit propagation of high-resolution images,'' in \emph{Proc.
  IEEE Conf. Comput. Vis. Pattern Vis. (CVPR)}, 2014, pp. 2854--2861.

\bibitem{endo2016deepprop}
Y.~Endo, S.~Iizuka, Y.~Kanamori, and J.~Mitani, ``Deepprop: Extracting deep
  features from a single image for edit propagation,'' in \emph{Comput. Graph.
  Forum}, 2016, pp. 189--201.

\bibitem{zhang2017real}
R.~Zhang, J.-Y. Zhu, P.~Isola, X.~Geng, A.~S. Lin, T.~Yu, and A.~A. Efros,
  ``Real-time user-guided image colorization with learned deep priors,''
  \emph{arXiv preprint arXiv:1705.02999}, 2017.

\bibitem{o2011color}
P.~O'Donovan, A.~Agarwala, and A.~Hertzmann, ``Color compatibility from large
  datasets,'' \emph{ACM Trans. Graph. (TOG)}, vol.~30, no.~4, pp. 1--12, 2011.

\bibitem{lin2013probabilistic}
S.~Lin, D.~Ritchie, M.~Fisher, and P.~Hanrahan, ``Probabilistic
  color-by-numbers: Suggesting pattern colorizations using factor graphs,''
  \emph{ACM Trans. Graph. (TOG)}, vol.~32, no.~4, pp. 1--12, 2013.

\bibitem{chang2015palette}
H.~Chang, O.~Fried, Y.~Liu, S.~DiVerdi, and A.~Finkelstein, ``Palette-based
  photo recoloring,'' \emph{ACM Trans. Graph. (TOG)}, vol.~34, no.~4, pp.
  139--1, 2015.

\bibitem{cho2017palettenet}
J.~Cho, S.~Yun, K.~Mu~Lee, and J.~Young~Choi, ``Palettenet: Image
  recolorization with given color palette,'' in \emph{Proc. Ieee Conf. Comput.
  Vis. Pattern Recognit. (CVPR) Workshops}, 2017, pp. 62--70.

\bibitem{ronneberger2015u}
O.~Ronneberger, P.~Fischer, and T.~Brox, ``U-net: Convolutional networks for
  biomedical image segmentation,'' in \emph{Int. Conf. Medical Image Comput.
  Comput.-Assisted Intervention}, 2015, pp. 234--241.

\bibitem{Gatys_2016_CVPR}
L.~A. Gatys, A.~S. Ecker, and M.~Bethge, ``Image style transfer using
  convolutional neural networks,'' in \emph{Proc. IEEE Conf. Comput. Vis.
  Pattern Recognit. (CVPR)}, June 2016.

\bibitem{li2017universal}
Y.~Li, C.~Fang, J.~Yang, Z.~Wang, X.~Lu, and M.-H. Yang, ``Universal style
  transfer via feature transforms,'' \emph{arXiv preprint arXiv:1705.08086},
  2017.

\bibitem{simonyan2014very}
K.~Simonyan and A.~Zisserman, ``Very deep convolutional networks for
  large-scale image recognition,'' \emph{arXiv preprint arXiv:1409.1556}, 2014.

\bibitem{beigpour2011object}
S.~Beigpour and J.~Van De~Weijer, ``Object recoloring based on intrinsic image
  estimation,'' in \emph{Int. Conf. Comput. Vis. (ICCV)}, 2012, pp. 327--334.

\bibitem{pitie2007linear}
F.~Piti{\'e} and A.~Kokaram, ``The linear monge-kantorovitch linear colour
  mapping for example-based colour transfer,'' in \emph{Proc. IET CVMP}, 2007,
  pp. 1--9.

\bibitem{grogan2015l2}
M.~Grogan, M.~Prasad, and R.~Dahyot, ``L2 registration for colour transfer,''
  in \emph{Proc. Eur. Signal Process. Conf.}, 2015, pp. 1--5.

\bibitem{he2016deep}
K.~He, X.~Zhang, S.~Ren, and J.~Sun, ``Deep residual learning for image
  recognition,'' in \emph{Proc. IEEE Conf. Comput. Vis. Pattern Recognit.
  (CVPR)}, 2016, pp. 770--778.

\bibitem{lin2014microsoft}
T.-Y. Lin, M.~Maire, S.~Belongie, J.~Hays, P.~Perona, D.~Ramanan,
  P.~Doll{\'a}r, and C.~L. Zitnick, ``Microsoft coco: Common objects in
  context,'' in \emph{Eur. Conf. Comput. Vis. (ECCV)}, 2014, pp. 740--755.

\bibitem{haralick1973textural}
R.~M. Haralick, K.~Shanmugam, and I.~H. Dinstein, ``Textural features for image
  classification,'' \emph{IEEE Trans. Syst., Man, Cybernetics}, no.~6, pp.
  610--621, 1973.

\bibitem{fridrich2012rich}
J.~Fridrich and J.~Kodovsky, ``Rich models for steganalysis of digital
  images,'' \emph{IEEE Trans. Inf. Forensics Security}, vol.~7, no.~3, pp.
  868--882, 2012.

\bibitem{pevny2010steganalysis}
T.~Pevny, P.~Bas, and J.~Fridrich, ``Steganalysis by subtractive pixel
  adjacency matrix,'' \emph{IEEE Trans. Inf. Forensics Security}, vol.~5,
  no.~2, pp. 215--224, 2010.

\bibitem{sullivan2006steganalysis}
K.~Sullivan, U.~Madhow, S.~Chandrasekaran, and B.~Manjunath, ``Steganalysis for
  markov cover data with applications to images,'' \emph{IEEE Trans. Inf.
  Forensics Security}, vol.~1, no.~2, pp. 275--287, 2006.

\bibitem{li2020identification}
H.~Li, B.~Li, S.~Tan, and J.~Huang, ``Identification of deep network generated
  images using disparities in color components,'' \emph{Signal Process.}, vol.
  174, p. 107616, 2020.

\bibitem{russakovsky2015imagenet}
O.~Russakovsky, J.~Deng, H.~Su, J.~Krause, S.~Satheesh, S.~Ma, Z.~Huang,
  A.~Karpathy, A.~Khosla, M.~Bernstein \emph{et~al.}, ``Imagenet large scale
  visual recognition challenge,'' \emph{Int. J. Comput. Vis.}, vol. 115, no.~3,
  pp. 211--252, 2015.

\bibitem{paszke2019pytorch}
A.~Paszke, S.~Gross, F.~Massa, A.~Lerer, J.~Bradbury, G.~Chanan, T.~Killeen,
  Z.~Lin, N.~Gimelshein, L.~Antiga \emph{et~al.}, ``Pytorch: An imperative
  style, high-performance deep learning library,'' \emph{Proc. Adv. Neural Inf.
  Process. Syst.}, vol.~32, pp. 8026--8037, 2019.

\bibitem{kingma2014adam}
D.~P. Kingma and J.~Ba, ``Adam: A method for stochastic optimization,''
  \emph{arXiv preprint arXiv:1412.6980}, 2014.

\bibitem{loshchilov2016sgdr}
I.~Loshchilov and F.~Hutter, ``Sgdr: Stochastic gradient descent with warm
  restarts,'' \emph{arXiv preprint arXiv:1608.03983}, 2016.

\bibitem{he2015delving}
K.~He, X.~Zhang, S.~Ren, and J.~Sun, ``Delving deep into rectifiers: Surpassing
  human-level performance on imagenet classification,'' in \emph{Proc. IEEE
  Int. Conf. Comput. Vis. (ICCV)}, 2015, pp. 1026--1034.

\bibitem{chattopadhay2018grad}
A.~Chattopadhay, A.~Sarkar, P.~Howlader, and V.~N. Balasubramanian,
  ``Grad-cam++: Generalized gradient-based visual explanations for deep
  convolutional networks,'' in \emph{IEEE Winter Conf. Appl. Comput. Vis.
  (WACV)}, 2018, pp. 839--847.

\bibitem{van2008visualizing}
L.~Van~der Maaten and G.~Hinton, ``Visualizing data using t-sne,'' \emph{J.
  Machine Learning Research}, vol.~9, no.~11, 2008.

\bibitem{chollet2017xception}
F.~Chollet, ``Xception: Deep learning with depthwise separable convolutions,''
  in \emph{Proc. IEEE Conf. Comput. Vis. Pattern Recognit. (CVPR)}, 2017, pp.
  1251--1258.

\bibitem{goljan2014rich}
M.~Goljan, J.~Fridrich, and R.~Cogranne, ``Rich model for steganalysis of color
  images,'' in \emph{IEEE Int. Workshop Inf. Forensics Security (WIFS)}, 2014,
  pp. 185--190.

\end{thebibliography}
\end{document}